\documentclass[authoryear,3p,times,twocolumn]{elsarticle}
\usepackage{amsthm}
\usepackage{url}
\usepackage{footnote}
\usepackage{amsmath}
\usepackage{algpseudocode}
\usepackage[ruled, vlined, linesnumbered, resetcount]{algorithm2e}
\usepackage{subfig}
\usepackage{placeins}
\usepackage[normalem]{ulem}
\usepackage{enumitem}
\usepackage{multirow}
\useunder{\uline}{\ul}{}

\usepackage{amssymb}
\usepackage{graphicx}

\usepackage{subfig}
\usepackage{multirow}

\journal{.}

\begin{document}

\begin{frontmatter}



\title{A \textit{Dynamic-Adversarial} Mining Approach to the Security of Machine Learning}


\author[tj]{Tegjyot Singh Sethi \corref{cor1}}
\ead{tegjyotsingh.sethi@louisville.edu}
\author[tj]{Mehmed Kantardzic}
\ead{mehmedkantardzic@louisville.edu}
\author[tj]{Lingyu Lyu}
\ead{l0lv0002@louisville.edu}
\author[jc]{Jiashun Chen}
\ead{1976139@163.com}


\address[tj]{Data Mining Lab, University of Louisville, Louisville, USA }
\address[jc]{Huai Hai Institute of Technology, China }

\cortext[cor1]{Corresponding author.}

\begin{abstract}
Operating in a dynamic real world environment requires a forward thinking and adversarial aware design for classifiers, beyond fitting the model to the training data. In such scenarios, it is necessary to make classifiers - a) harder to evade, b) easier to detect changes in the data distribution over time, and c) be able to retrain and recover from model degradation. While most works in the security of machine learning has concentrated on the evasion resistance (a) problem, there is little work in the areas of reacting to attacks (b and c). Additionally, while streaming data research concentrates on the ability to react to changes to the data distribution, they often take an adversarial agnostic view of the security problem. This makes them vulnerable to adversarial activity, which is aimed towards evading the concept drift detection mechanism itself. In this paper, we analyze the security of machine learning, from a dynamic and adversarial aware perspective. The existing techniques of Restrictive one class classifier models, Complex learning models and Randomization based ensembles, are shown to be myopic as they approach security as a static task. These methodologies are ill suited for a dynamic environment, as they leak excessive information to an adversary, who can subsequently launch attacks which are indistinguishable from the benign data. Based on empirical vulnerability analysis against a sophisticated adversary, a novel feature importance hiding approach for classifier design, is proposed. The proposed design ensures that future attacks on classifiers can be detected and recovered from. The proposed work presents motivation, by serving as a blueprint, for future work in the area of \textit{Dynamic-Adversarial} mining, which combines lessons learned from Streaming data mining, Adversarial learning and Cybersecurity. 
\end{abstract}
 
\begin{keyword}
Adversarial machine learning \sep Streaming data  \sep Concept drift \sep Ensemble \sep Feature information hiding \sep Evasion
\end{keyword}

\end{frontmatter}


\section{Introduction}
\label{sec:introduction}

\begin{figure}[t]
  \centering
  \includegraphics[width=0.9\linewidth]{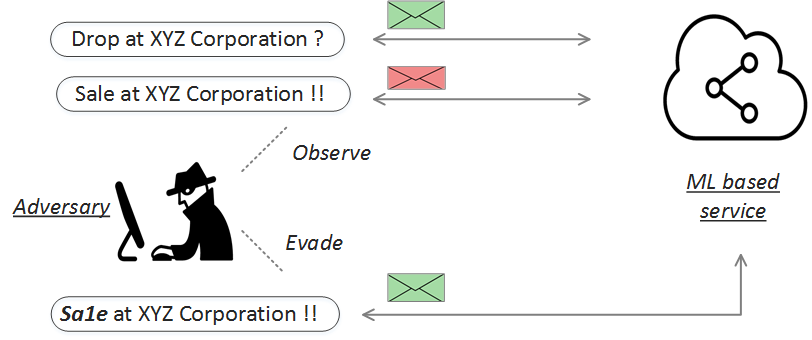}
   \caption{Illustration of exploratory attacks on a machine learning based spam filtering system.}
  \label{fig:evasion1}
\end{figure}

The machine learning gold rush has led to a proliferation of predictive models, being used at the core of several cybersecurity applications. Machine learning allows for scalable and swift interpolation from large quantities of data, and also provides generalize-ability to improve longevity of the security mechanisms. However, the increased exuberance in the application of machine learning has led to the overlooking of its security vulnerabilities. Recent works on adversarial machine learning \citep{tsethi2016,papernot2017practical,papernot2016limitations,tramer2016stealing}, have demonstrated that deployed classification systems are vulnerable to adversarial perturbations at test time, which cause the models to degrade over time. Such attacks on the integrity of machine learning systems, are termed as exploratory attacks \citep{barreno2006can}, as they are caused at test time, by an adversary learning and evading the characteristics of the deployed classifier model. An example of such an attack is shown in Figure~\ref{fig:evasion1}, where an adversary learns the behavior of the spam detection system, and then crafts emails so as to avoid being flagged. Similar attacks have been shown to be possible on computer vision systems \citep{papernot2016limitations}, speech recognition systems \citep{carlini2016hidden} and also on natural language processing systems \citep{hosseini2017deceiving}. 

\begin{figure}[t]
\centering
\subfloat[\textit{Proactive security} aims at increase time and effort needed ($\delta_P$) to degrade a model ]{\includegraphics[width=0.44\linewidth]{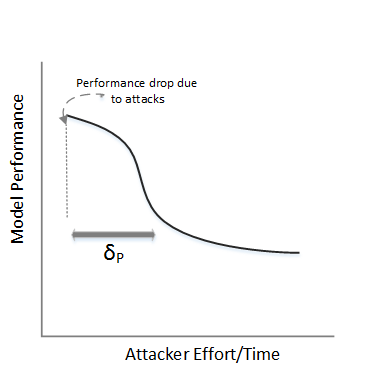}}
\hspace{1em}
\subfloat[\textit{Reactive security} aims at reduce the time taken to detect attacks and fix them ($\delta_R$)]{\includegraphics[width=0.44\linewidth]{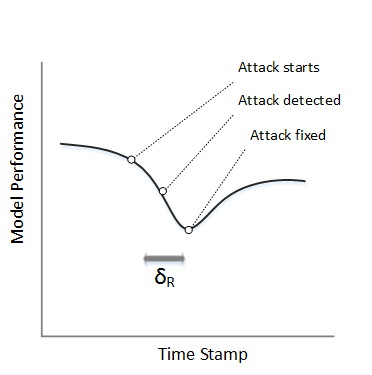}}
\caption{Goals of \textit{Proactive} and \textit{Reactive} Security.}
\label{fig:security_paradigms}
\end{figure}

Owing to the pressing need for secure machine learning systems, there has been considerable recent attention towards improving the resilience of such systems \citep{papernot2016towards,wang2015robust,smutz2016tree}. However, the entire research effort in the security of machine learning is split into two schools of thought: a) \textit{Proactive security} measures, and b) \textit{Reactive security} measures. \textit{Proactive security} is aimed towards anticipating adversary strategies, and making systems resistant to possible attacks \citep{hardt2016strategic,wang2015robust,biggio2010multiple,stevens2013hardness,xu2014comparing,colbaugh2012predictive,vorobeychik2014optimal}. These methods focus on \textit{prevention}, with the aim of delaying an attack till the maximum possible extent. As shown in Figure~\ref{fig:security_paradigms}) a), these defensive measures aim at increasing the attackers effort $\delta_P$, which is needed to degrade the target system's performance (evade the system). Work on proactive security views the attack-defense problem as a static one, with the intention of bolstering defenses before deploying the system. 

While works on proactive security emphasize on the prevention problem, \textit{Reactive security} focuses on the problem of fixing the system after it is attacked \citep{barth2012learning,henke2015analysis}. Reactive security emphasizes on the \textit{detection} component of the attack-defense cycle. These approaches aim at being able to swiftly detect attacks and fix the system, with limited human supervision, to make the system available and effective again. As shown in Figure~\ref{fig:security_paradigms}) b), these measures aim at reducing the delay $\delta_R$, which is the time taken by the system to detect degradation and fix (retrain) the model. Works in the area of concept drift detection and adaptation are suitable for reactive security, as they do not make any explicit assumption about the type of change and aim at maintaining high model performance, over time \citep{minku2012ddd,brzezinski2014reacting,gama2014survey}. However, hitherto, the work in the domain of concept drift research has followed a domain agnostic approach, where any changes to the data distribution is regarded equally, without any adversarial awareness. 

The works on \textit{Proactive security} and \textit{Reactive security}, approach the challenge of security from different perspectives. An analogy to explain the two is: Consider the task of securing a precious artifact. \textit{Proactive security} aims at securing the vault in which the artifact is kept- by making thicker walls, by setting up barricades, chains and locks, and by keeping track of suspects who have a history of theft. On the other hand, \textit{Reactive security} would concentrate its efforts on setting up surveillance and alarm systems, so as to be able to apprehend the thief easily, if at all a larceny is attempted. There is a need for a holistic approach, which combines the benefits of both these classes of security measures.

\begin{figure}[t]
\centering
\subfloat[Impact of using a restrictive defender model.]{\includegraphics[width=0.95\linewidth]{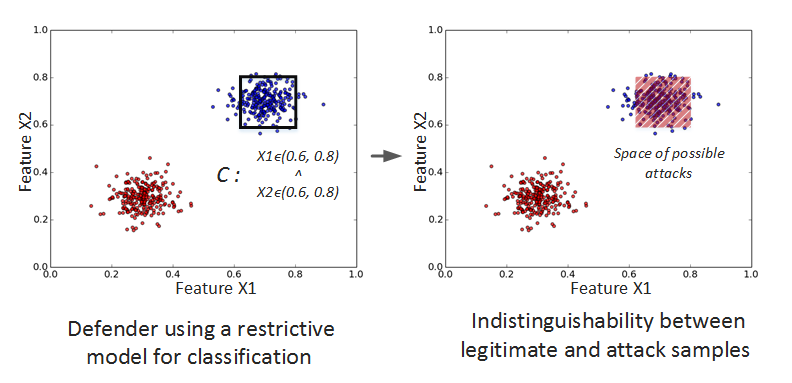}}
\\
\subfloat[Impact of using a simple defender model.]{\includegraphics[width=0.95\linewidth]{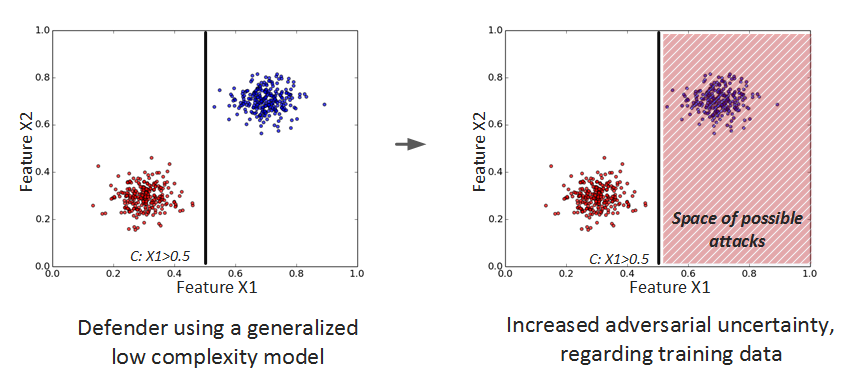}}

\caption{Impact of using a restrictive defender model (top) and that of using a more generalized defender model (bottom). In case if restrictive models, attacks are harder to carry out, but will lead to inseparability from benign traffic.}
\label{fig:model_informativeness1}
\end{figure}

 Since the adversary is in a never-ending arms race with the defender, a dynamic and adversarial aware approach to security is needed. The need for a holistic solution, is illustrated using the example of a 2D synthetic dataset in Figure~\ref{fig:model_informativeness1} a). A classifier model $C$ is trained on the space of \textit{Legitimate} class training samples (blue). The model $C$ is seen to be restrictive, as it allows only a narrow margin of samples to be admitted into the system as benign samples. This strategy is considered to provide better security from a static perspective, as it would require an adversary to craft samples in a small and exact range of feature values. However, looking at the problem from a dynamic perspective, it is seen that the space of possible attacks is now highly overlapped with the training data. As such, once an attack starts, the adversary has enough information about the feature space, so as to become indistinguishable from benign traffic entering the system. These attacks are harder to detect and also harder to recover from, due to the inseparability of samples. On the other hand, the simple design of Figure~\ref{fig:model_informativeness1} b) (simple as it reduces the number of features used in the model $C$), has a larger space of possible attacks, making it easier to evade. However, the increased space means that an adversary evading $C$ will not be able to completely reverse engineer the location and characteristics of the \textit{Legitimate} training samples, due to the added uncertainty provided by the large space of possible attacks. 

The illustrative example of Figure~\ref{fig:model_informativeness1} a) and Figure~\ref{fig:model_informativeness1} b), demonstrate the need to develop countermeasures from a \textit{Dynamic - Adversarial} perspective. In such an environment, delaying the onset of attacks, detecting attacks and recovering from them are all equally important, for the continued usage of the classification system. In this paper, the impact of classifier design strategies on the severity of attacks launched at test time, is presented. The existing ideas of \textit{Randomness} \citep{alabdulmohsin2014adding,colbaugh2012predictability} and \textit{Complex learning} \citep{rndic2014practical,wang2015robust,biggio2010multipleattack}, for securing classifiers, are evaluated from a dynamic perspective.  Also, the ability to take preemptive measures, to ensure defender's leverage in a dynamic environment, is evaluated. In particular, the ability to hide feature importance is demonstrated, as a proactive measure to enable efficient reactive security. To the best of our knowledge, this is the first work which combines both adversarial and dynamic aspects of the security of machine learning. The main contributions of this paper are as follows:

\begin{itemize}
\item Analyzing the impact of classifier design strategies, in the training phase, on the dynamic capabilities of an adversary, at test time. 
\item Quantifying the impact of attack severity using the Adversarial Margin Density (AMD) metric, for determining the ability to detect and recover from attacks. 
\item Analyzing vulnerabilities of popularly used Restrictive, Complex learning and Randomization based approaches, in a dynamic environment. 
\item Proposing a counter-intuitive and novel feature hiding approach, to ensure long term defender leverage, for better attack detection and model retrain-ability. This proposed approach serves as a blueprint for future works in the domain of \textit{Dynamic-Adversarial} mining.
\item Providing background and motivation for future work, for a new interdisciplinary area of study - \textit{Dynamic-Adversarial} mining, which combines Streaming data, Machine learning, Cybersecurity and Adversarial learning, for a holistic approach to the security of machine learning. 
\end{itemize}

The rest of the paper is organized as follows: Section~\ref{sec:lr} presents a survey of work in the area of security of machine learning. Section~\ref{sec:pm} presents the analysis methodology for evaluating effects of different classifier design strategies, on the adversarial activities at test time.  Section~\ref{sec:experimentation_5} presents experimental evaluation of popular classifier design strategies in the domain of security: Restrictive one class classifiers, Complex learning based ensemble methods and Randomization based classifiers, are evaluated. Based on the analysis, a novel feature hiding approach is proposed and evaluated in Section~\ref{sec:hidden}. Ideas and motivation for future work in the domain of \textit{Dynamic-Adversarial} mining, are presented in Section~\ref{sec:da}. Conclusion and avenues for further extensions of the work are presented in Section~\ref{sec:conclusion}.

\section{Related work on security of machine learning}
\label{sec:lr}

In this section, related work in the domain of machine learning security is discussed. Section~\ref{sec:lr_attacks} presents work in the area of adversarial manipulation at test time, which affects the performance of machine learning based systems. Proactive/Static security measures, developed in the domain of adversarial machine learning, are discussed in Section~\ref{sec:lr_proactive}. Methodologies which advocate a reactive/dynamic approach to dealing with attacks are discussed in Section~\ref{sec:lr_reactive}. The need for a holistic approach, combining dynamic and adversarial thinking, is emphasized.

 \subsection{Exploratory attacks on classification based systems}
\label{sec:lr_attacks}

Machine learning systems deployed in the real world are vulnerable to exploratory attacks, which aim to degrade the learned model, over time. Exploratory attacks are launched by an adversary, by first learning about the characteristics of the system through carefully crafted probes, and then morphing suspicious samples to cause evasion at test time \citep{biggio2014pattern,tramer2016stealing,biggio2013evasion,papernot2016transferability,biggio2014security,tsethi2016}. Such attacks are commonplace and difficult to avoid, as they rely on the same black box access to the systems, that a benign user is entitled to. These attacks were shown to affect a wide variety of classifier types in \citep{papernot2016transferability}, and were also shown to be potent towards \textit{ML-as-a-service} (such as Amazon AWS Machine Learning\footnote{\url{https://aws.amazon.com/machine-learning/}} and Google Cloud Platform\footnote{\url{cloud.google.com/machine-learning}}),  which provide APIs for accessing predictive analytics as a service \citep{sethi2017data}.

 \begin{figure}[t]
  \centering
  \includegraphics[width=0.95\linewidth]{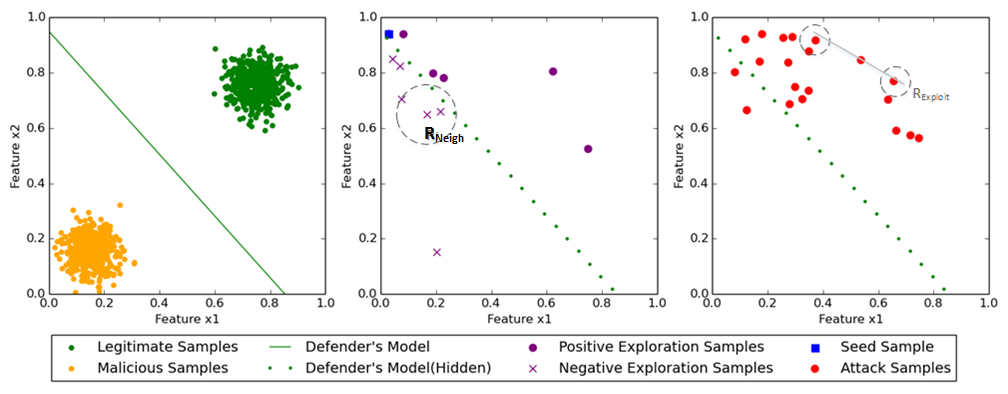}
   \caption{Illustration of AP attacks. \textit{(Left - Right)}: The defender's model from its training data. The Exploration phase depicting the seed (blue) and the anchor points samples (purple). The Exploitation attack phase samples (red) generated based on the anchor points.\citep{tsethi2016}}
  \label{fig:ap}
\end{figure}

One such black box, data driven, and algorithmic attack strategy, was proposed in \citep{tsethi2016}. The work in \citep{tsethi2016}, proposed the Anchor Points (AP) attack strategy, which uses an \textit{exploration-exploitation} framework, to perform evasion on classifiers, as shown in Figure~\ref{fig:ap}. In these attacks, the adversary starts with exploration of the deployed classifier model(green), by using crafted probing samples. These samples are submitted to the system, via remote API calls, and the resulting feedback of \textit{Accept/Reject} is observed. The AP attack strategy then leverages the samples which lead to successful evasion, to generate additional attack samples (red) via exploitation. These attack samples will cause the performance of the defender to drop, ultimately leaving it unusable. The AP attack strategy provides a domain independent, classifier independent, and simplistic approach to simulate exploratory attacks and to evaluate the impact of various security design decisions, on the performance of the classifier. In this paper, the AP attack strategy will be extended and used, as a tool for assessing and comparing the efficacy of various security mechanisms.

\subsection{Proactive/Static approaches for security of machine learning}
\label{sec:lr_proactive}

Works on proactive security, approach the problem from a static perspective. Robustness is incorporated into the model's design, when it is trained from the observed data, and the system is assumed to be resilient against future evasion attacks. These techniques are popular in the adversarial machine learning research domain, and the available work can be broadly classified into the following two categories: a) Complex learning methods and b) Disinformation methods. These categories of attacks are discussed in the following sections.

\subsubsection{Security based on learning of complex models}
\label{sec:clm}

These methods increase the complexity of the learned model, to make it harder for an attacker to effectively reverse engineer it with a limited number of probes. The complexity is increased either by using difficult to mimic features in the classification model \citep{rndic2014practical}, or by balancing weights across features, such that the resulting classification models are robust \citep{rndic2014practical,wang2015robust,biggio2010multipleattack}. Such robust classifiers require the adversary to reverse engineer and mimic a large number of features, thereby increasing its efforts and cost, to successfully evade the classifier. A practical challenge that these methods face is the balancing of complexity with overfitting, as overfitting can lead to poor generalization performance and can also make the system vulnerable to training time (causative) attacks \citep{barreno2010security}. 
 
Complexity of the models can be increased systematically, by distributing weights among informative features, to make a classifier robust \citep{rndic2014practical}. In \citep{kolcz2009feature}, feature reweighing was used by weighting every feature inversely based on their importance. This made the classification boundary dependent on a larger number of features, thereby requiring attacks to spend more effort in trying to evade the classifier. In \citep{zhang2015adversarial}, the task of selecting a reduced feature subset in an adversarial environment was presented. Classification security was introduced as a regularization term, in the learning objective function of the defender's model. This term was optimized along with the classifier's generalization capability, during the feature selection process, making it more secure against evasion attacks on reduced feature spaces. Robustness to random feature deletion at test time, caused by noise or by adversarial reverse engineering, was introduced in \citep{globerson2006nightmare}. It was shown that the proposed methodology was resistant to changes caused by random feature deletion and also to, a reasonable extent, those caused by deletion based on feature importance. In \citep{colbaugh2012predictive} the predictive defense algorithm was proposed, which adds regularization terms in the min-max formulation of the objective function, to model adversary actions in aggressively reduced feature spaces. The objective function optimized in \citep{colbaugh2012predictive}, is given by Equation~\ref{eqn:moving_target}.

\begin{equation}
\min _{ w }{ \max _{ a }{ \left[ -\alpha { \left\| a \right\|  }^{ 3 }+\beta { \left\| w \right\|  }^{ 3 }\quad +\sum _{ i }{ loss({ y }_{ i },{ w }^{ T }({ x }_{ i }+a)) }  \right]  }  } 
\label{eqn:moving_target}
\end{equation}

Where, the loss function represent misclassification rate and \textit{w} represents the learned model of the defender. The attacker attempts to circumvent the defender by using a linear transform vector $a\in R^d$. The terms $\alpha$ and $\beta$ denote the regularization imposed on attackers and defenders, respectively. The term $\alpha$ embodies the effort of an attacker, which increases based on the distance from existing samples. The term $\beta$ denotes the defenders need to avoid overfitting to the data. By incorporating both regularization terms into the optimization function, this method makes it harder for an adversary to apply a simple linear transformation on a few features of malicious samples, to get them classified as \textit{Legitimate} by the defender model. The use of this game theoretic formulation to the problem was shown to outperform a naive Bayes classifier, used as gold standard, for the task of spam classification with reduced feature space.

While the above methods rely on adding robustness to the training of a single classifier, it was shown in \citep{biggio2008adversarial,biggio2010multiple,biggio2010multipleattack} that combining classifiers trained on different subsets of the data is more suitable to the task of security. Multiple Classifier Systems (MCS)\citep{wozniak2014survey} allow classifiers trained on different subsets of the feature space to be combined in a natural way, to provide an overall robust classification prediction. In \citep{biggio2010multiple}, instance bagging and random subspace methods were analyzed, as two MCS approaches, for securing against evasion attacks. It was found that both approaches make the evasion problem harder for the adversary. In particular, random subspace models are more suitable when a very small number of features exhibit high discriminating capability, whereas bagging works better when the weights are already evenly distributed among the features. These methods essentially have the effect of averaging over the feature weights, as models trained on different views of the problem space are aggregated to present the final result\citep{biggio2010multipleattack}. Heterogeneous combination of one class and binary classifiers, to produce an evasion resistant model, has also shown to balance high accuracy with improved security, in works of \citep{biggio2015one}.

\subsubsection{Security based on disinformation}
\label{sec:randomization}

These methods rely on the concept of \textit{`Security by obscurity'}, and they are aimed at confusing the adversary about the internal state of the system \citep{biggio2014pattern}. The inability of the adversary to accurately ascertain the internal state would then result in increased delay/effort for performing the attack. Popular principles for applying disinformation based security are: Randomization \citep{huang2011adversarial}, Use of honeypots \citep{rowe2007defending}, Information hiding \citep{abramson2015toward} and Obfuscation \citep{barreno2006can}. An example obfuscation technique used in case of a classifier filter, is one where the classifier would output correct prediction with a probability of 0.8, in an attempt to sacrifice accuracy to provide security. Ideally, hiding all information about the feature space and classification model could result in complete security. However, this goes against the Kerchkoff's principles of information security \citep{kerckhoffs1883cryptographie,mrdovic2008kerckhoffs}, which states that security should not rely overly on obscurity and that one should always assume that the adversary knows the system. 

When disinformation relies on obfuscation, the amount of information made available to the adversary is not limited, but obtaining the information is made harder \citep{xu2014comparing}. Randomization is a popular approach to ensure that information presented is garbled. A simple strategy for introducing randomization into the classification process was proposed in \citep{barreno2006can}, where it was suggested that randomness be introduced in placing the boundary of the classifier. This was made possible by using the confidence estimate given by a probabilistic classifier, to randomly pick the class labels. While this increases the evasion efforts necessary, it leads to a drop in accuracy in non adversarial environments. A detailed analysis of this tradeoff was presented in \citep{alabdulmohsin2014adding}, where a family of robust SVM models were learned and used to increase reverse engineering effort. 

The work in \citep{alabdulmohsin2014adding} showed that choosing a single classifier is a suboptimal strategy, and that drawing classifiers from a distribution with large variance can improve resistance to reverse engineering, at little cost to the generalization performance of the model. This idea has also been used in multiple classifier systems, where a random classifier is selected at any given time, from a bag of trained classifiers, to perform the classification. The moving target approach of \citep{colbaugh2012predictability}, divides features into \textit{K} subsets and trains one classifier on each of these subsets. One of these classifiers is chosen, according to a scheduling policy, to perform the classification of an incoming sample \textit{x}. It was shown that using a random uniform scheduling policy provides the best protection against an active adversary. This was formally proven in \citep{vorobeychik2014optimal}, where it was shown that the optimal randomization strategy is to select classifiers randomly with equal probabilities. In case of targeted attacks, selecting uniformly was shown to be the best strategy, whereas no randomization was the optimal strategy in the face of indiscriminate attacks. Although the ideas of \citep{colbaugh2012predictability} and \citep{vorobeychik2014optimal} are applicable to multiple classifier systems, experimental results were shown only on a set of two separately trained classifiers. Obtaining multiple sets of classifiers, with high accuracy, was not discussed. The work of \citep{biggio2008adversarial} uses the multiple classification system formulation, by maintaining a large set of classifier models and with randomization being performed, by assigning a different weighting scheme to the individual classifier's prediction, for every iteration. Experimentation was performed on the Spamassassin dataset, by pre-selecting 100 weighting schemes and allowing the classifier to choose any one of them with equal probability, over different iterations. This resulted in increased hardness of evasion of the classifier under simulated attacks. 

While randomization is a popular method for incorporating obscurity against evasion attacks, other methods have been suggested \citep{biggio2014pattern}. Limiting the feedback or providing incorrect feedback, to combat probing attacks have been suggested in \citep{barreno2006can}. Use of honeypots is another promising direction, wherein systems are designed for the sole purpose of being attacked and thereby providing information about attackers \citep{rowe2007defending}. Use of social honeypots was suggested in \citep{lee2010uncovering}, as a means of collecting adversarial samples about social spam. Bots which behave as humans are deployed in the social cyber-space, specifically tested on Myspace and Twitter data in \citep{lee2010uncovering}, where they collect information about potential spammers.

\subsection{Reactive/Dynamic approaches for security of machine learning}
\label{sec:lr_reactive}

\begin{figure}[t]
  \centering
  \includegraphics[width=0.5\linewidth]{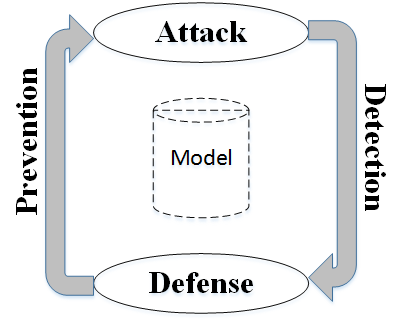}
   \caption{The Attack-Defense cycle between adversaries and system designers (i.e., defenders). }
  \label{fig:cycle}
\end{figure}

Adversarial learning is a cyclic process, as described in Figure~\ref{fig:cycle} and in \citep{stein2011facebook}. Attacks are a question of \textit{when} and not \textit{if}. Proactive measures of security delay the occurrence of attack, but eventually every system is vulnerable to attack. These measures do not provide any direction or solution for dealing with attacks, and for fixing the system for future use after it has been attacked. They are static one shot solutions, which drop the ball as soon as a system is compromised. Reactive system on the other hand, can deal with attacks after their onset and aim to fix the system as soon as possible. Work in \citep{barth2012learning} shows that, in the absence of prior information about adversaries, reactive system performs as good as a proactive security setting and is a suggested security infrastructure for industrial scale systems. Reactive systems can also ensure that attacks are recognized and alerted, to take appropriate counter measures. 

Reactive methods have primarily been studied in the domain of streaming data mining, where changes to the data, called concept drift, are detected and the system is adapted to maintain performance over time \citep{vzliobaite2010learning}. In these techniques, the system is agnostic to the cause of change, and treats any degradation in its performance in a similar way. As such, they are reactive but not necessarily adversary aware \citep{dalvi2004adversarial}. The work in \citep{gama2014survey} summarizes the major frameworks, methodologies and algorithms developed for handling concept drift. A trigger based concept drift handling approach \citep{ditzler2015learning} consists of two important components: a) A drift detection module, and b) A drift adaptation module. The drift detection module is responsible for detecting changes in the data and to signal further evaluation or adaptation based on the detected changes. The adaptation module then launches retraining of the system, by collecting new labeled data and updating the model. Ensemble techniques are a popular choice for drift handling systems, as they allow modular retraining of the system \citep{sethi2016grid}. Ensemble methods have been proposed for drift detection \citep{kuncheva2008classifier} and for learning with concept drift \citep{brzezinski2014reacting,minku2012ddd}.

At the confluence of concept drift and adversarial learning, are the approaches of adversarial drift \citep{kantchelian2013approaches} and adversarial active learning \citep{miller2014adversarial}. In \citep{kantchelian2013approaches}, concept drift caused as a result of attacks, was discussed. The need for a responsive system, with the integration of traditional blacklists and whitelists, alongside an ensemble of classifiers trained for every class of malicious activity, was proposed. Isolation of malicious campaign and techniques to ensure zero training error, while still maintaining generalization, were extensions suggested for traditional classification systems when used in an adversarial environment. The vulnerabilities of the labeling process were analyzed in \citep{miller2014adversarial}, which introduces the concept of adversarial active learning. The Security oriented Active Learning Test bed (SALT) was proposed in \citep{miller2014adversarial}, to ensure effective drift management, human integration and querying strategies, in an adversarial environment.  Use of concept drift tracking within malware families was analyzed in \citep{singh2012tracking}, to show the temporal nature of adversarial samples. Metafeatures (higher order difficult to evade) were suggested, to detect malicious activity. Use of an ensemble of classifier to detect spam emails was suggested in \citep{chinavle2009ensembles}, where mutual agreement of pairs of classifiers in the ensemble was tracked and concept drift was detected if the agreement drops. In the event of a drift detection, poorly performing pairs of classifiers are selected to be retrained.  Extension of this work was proposed in \citep{smutz2016tree}, where the entire ensemble's disagreement score distribution was tracked. A sudden increase in the overall disagreement was used to indicate an attack on the system, without using external labeled data. Here, feature bagging \citep{bry} was found to be an effective ensemble strategy to detect evasion attacks, such as mimicry and reverse mimicry attacks, on the task of classifying malicious pdf documents. These works provide initial ideas for using machine learning in an adversarial environment, where the data is streaming and the attacks-defense is a cyclic never-ending process (Figure~\ref{fig:cycle}).
 
The reactive approaches have been well studied towards an online never ending learning scheme. However, the current methods for reactive security do not explicitly consider an adversarial environment, even though a dynamic environment is considered. In adversarial domains, concept drift (attacks) are a function of the deployed model itself, as reverse engineering is based on what model is learned in the first place \citep{barreno2006can}. This information can be used to design more suited reactive systems, in adversarial domains. A combination of proactive and reactive approaches are necessary when dealing with such domains, as decisions made in the design phase could ultimately make future steps down the pipeline easier. Further research in this area will need a comprehensive look at the learning process, with adversarial effects considered at every step of the process, towards an adversarial aware dynamic learning system. This will require newer metrics of measuring system performance, beyond accuracy and f-measure, towards metrics such as recovery time and data separability \citep{mthembu2008note}, which are necessary for subsequent cycles of the process. In this paper, we look at the \textit{Dynamic-Adversarial} nature of the security of machine learning. To this end, we analyze the effect of various design strategies, on the long term security of a system.

\section{Adversarial Uncertainty - On the ability to react to attacks in \textit{Dynamic - Adversarial} environments}
\label{sec:pm}

Understanding the impact of attacks and defense in a cyclic environment, where a system is attacked and needs to recover from it periodically, requires a new perspective on the problem and the metrics used for evaluating the different design strategies. The metric of attack evasion rate \citep{lowd2005adversarial,tsethi2016} is sufficient for evaluating robustness of models to the onset of exploratory attacks,  but it does not provide intuition about the effectiveness of strategies which are designed for reacting to attacks in a dynamic environment. In this case, we need a metric which can convey information about the ability to react to attacks and continue operations, following the onset of attacks. In this section, the idea of \textit{Adversarial Uncertainty} is introduced, to understand the impact of different classifier design strategies on the ability to react to attacks, once they start to impact a system's performance. The motivation behind adversarial uncertainty is presented in Section~\ref{sec:5_motivation}. The impact of the defender's classifier design on adversarial uncertainty, is presented using an illustrative example of binary  feature valued data in Section~\ref{sec:classifier_adversarial}. A heuristic method to measure adversarial uncertainty is presented in Section~\ref{sec:amd}, by means of introducing the \textit{Adversarial Margin Density} (AMD) measure. The ability of adversaries to utilize the available probe-able information, from the defender's black box model, in order to launch high confidence attacks, is demonstrated in Section~\ref{sec:high_confidence} and Section~\ref{sec:ap_hc}. This is done to analyze the impact of a determined adversary, and to be able to design secure classifiers to counter them.

\subsection{Motivation}
\label{sec:5_motivation}

The defender extracts information about the predictive problem, by analyzing and learning from the training data. The adversary on the other hand, obtains its information by probing the deployed black box model of the defender. As such, there is a gap between the information held by the two players. This is not a significant concern when the adversary is aiming to only evade the deployed model, as complete understanding of the feature space is often not needed to evade most robust classifier models. However, this information deficit/gap is necessary to evaluate the impact an adversary can have on the reactive capabilities of a defender. Adversarial uncertainty refers to the uncertainty on the part of the adversary, due to the unavailability of the original training data. 

To understand the impact of adversarial uncertainty, consider the following toy example: a 2-dimensional binary dataset, where the sample \textit{L(X1=1, X2=1)} represents the \textit{Legitimate} class training sample and \textit{M(X1=0, X2=0)} represents the \textit{Malicious} class samples. A model trained as $C:X1 \vee X2$ provides generalization capabilities, by allowing (0,1) and (1,0) to be considered as legitimate samples at test time. In this case, an adversary looking to evade $C$, can pick one of the following three attack samples  at test time: (1,0), (0,1), (1,1). While any of these samples will lead to a successful evasion on the adversary's part, only one is truly devastating for a reactive system. The adversarial sample (1,1) will make the defender's model ineffective, as it completely mimics the training sample $L$. However, in this case, the probability of selecting this sample is 1/3, as an adversary is not certain about the exact impact of the features $X1$ and $X2$, on $C$. This uncertainty on the part of the adversary is referred to as adversarial uncertainty. In this example, adversarial uncertainty will enable the defender to recover from attacks 2/3 times, as the attack sample (1,0) can be thwarted by an updated model $C':X2$, and the sample (0,1) can be thwarted by the model $C':X1$. 

\begin{figure}[t]
  \centering
  \includegraphics[width=0.95\linewidth]{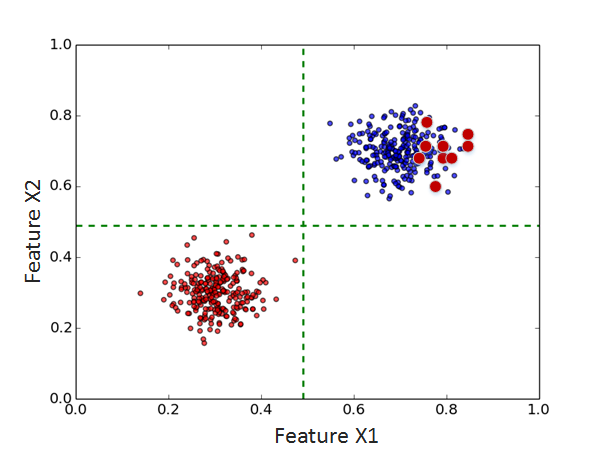}
   \caption{Illustration of data nullification attacks on the space of legitimate samples (blue).}
  \label{fig:data_nullification_numeric}
\end{figure}

The attack sample (1,1) is an example of a data nullification attack \citep{kantchelian2013approaches}. Data nullification leaves the defender unable to use the same training data, to continue operating in a dynamic environment. An illustration of data nullification attacks on numerical data spaces is shown in Figure~\ref{fig:data_nullification_numeric}. In this figure, the attacks samples generated, overlap with the original space of legitimate data samples. As such, the defender will be unable to train a high quality classification model, which can effectively separate the two class of samples, while using the existing set of features. Recovering from data nullification attacks will require the defender to generate additional features and collect new training data, both of which are time taking and expensive operations. Also, such attacks will be harder to detect via unsupervised techniques \citep{sethi2017reliable}, because the legitimate and the attack samples have a high degree of similarity and proximity, in the feature space. It is therefore required to ward of data nullification attacks, to be able to effectively detect and recover from attacks. Adversarial uncertainty provides a heuristic indication of the inability of the adversary, to deduce with confidence the exact impact of the various features on the prediction problem. A high adversarial uncertainty will lead to a lower probability of a data nullification attack.

From a dynamic data perspective, the ability to detect attacks and being able to retrain the classifier, are important characteristics of the system. This is possible only if the original legitimate training data is not corrupted by an adversary at test time (i.e., data nullification attacks are prevented). Data nullification attacks are possible if an adversary is able to simultaneously and successfully reverse engineer the impact of the entire feature set, on the defender's classification model. This is a result of an adversary's confidence in the impact of the different features on the prediction task, which it obtains via exploration of the deployed black box model.  A highly confident adversary can not only evade the deployed classifier, but can also avoid detection by unsupervised distribution tracking methodologies \citep{sethi2017reliable}. Thus, it is essential to evaluate the impact of the various defender strategies, on their ability to ensure that they do not leak excessive information to an adversary (i.e., their ability to ensure high adversarial uncertainty).

\subsection{Impact of classifier design on adversarial uncertainty}
\label{sec:classifier_adversarial}

Adversarial activity and capabilities are a function of the deployed black box classifier, which is being evaded. The adversary's perception of the prediction space is directed by the feedback on probes submitted to the defender's model. As such, the defender's classifier design has a certain degree of control in defining the range of attacks, that can be launched against it. Classifier design strategies range from restrictive one class classifiers \citep{salem2008survey,onoda2012analysis}, to robust majority voted classifier \citep{globerson2006nightmare, liu2012efficient}, and randomization based feature bagged ensembles \citep{colbaugh2012predictability, biggio2008adversarial}. From an adversarial perspective, the selection of classifier learning strategies can have a significant impact on adversarial uncertainty, and on the hardness of evasion. The impact of popular classifier design strategies,  based on research directions in security of machine learning, is illustrated here using an example of a $N$-dimensional binary feature space. We consider the training dataset to be comprised of one \textit{Legitimate} class sample \textit{L(1,1,...(N features),1)} and one \textit{Malicious} class sample \textit{M(0,0, ....(N features),0)}. As such, the features 1 to $N$ are all informative in discriminating between the two classes of samples. The effective usage of this orthogonal information, leads to the various design strategies, as shown in Table~\ref{tbl:classifier_designs}. Since this is meant to be an illustrative example, we consider the impact of these designs against a random probing based attack strategy, where the attacker tries different permutations of the $N$ binary variables, to understand the behavior of the black box classifier. 

\begin{table*}[t]
\centering
\caption{Impact of classifier design on evasion probability and adversarial certainty.}
\label{tbl:classifier_designs}
\scalebox{0.75}{
\begin{tabular}{|l|l|l|l|}
\hline
\begin{tabular}[c]{@{}l@{}}Classifier \\ Model (C)\end{tabular} & \begin{tabular}[c]{@{}l@{}}Model \\ representation\end{tabular} & \begin{tabular}[c]{@{}l@{}}Evasion \\ probability\end{tabular} & \begin{tabular}[c]{@{}l@{}}Adversarial \\ Certainty\end{tabular} \\ \hline
\begin{tabular}[c]{@{}l@{}}Simple model\\ (e.g., C4.5 Decision tree)\end{tabular} & $X_i=\quad 1$ & $2^{N-1}/2^N=\quad 1/2$ & $1/2^{N-1}$ \\ \hline
\begin{tabular}[c]{@{}l@{}}Complex model - \\ One class Legitimate\end{tabular} & $\bigwedge _{ i\in N }{ X_{ i } } =\quad 1$ & $1/2^N$ & 1/1=1  \\ \hline
\begin{tabular}[c]{@{}l@{}}Complex model - \\ One class Malicious\end{tabular} &  $\bigvee _{ i\in N }{ X_{ i } } =\quad 1$ & $(2^N-1)/2^N$ & $1/(2^N-1)$ \\ \hline
\begin{tabular}[c]{@{}l@{}}Complex model - Feature bagged \\ robust model (e.g., random subspace\\ majority voted ensemble)\end{tabular} & $\bigvee _{ s\subset N }{ \bigwedge _{ i\in s }{ X_{ i } } =\quad 1 } $ & $\sum _{ i=\frac { N }{ 2 } +1 }^{ N }{ \left( \begin{matrix} N \\ i \end{matrix} \right)  }/2^{N} =\quad 1/2^*$ & $1/2^{N-1}$ \\ \hline
\begin{tabular}[c]{@{}l@{}}Randomized model - Feature bagged \\ robust model with majority voting and \\ random selection\end{tabular} & $ \bigwedge _{ i\in s }{ X_{ i } } =\quad 1 ; \quad s\subset  N$ & $1/2^{N/2}$ & $1/2^{(N/2)}$  \\ \hline
\end{tabular}
}
$^*\sum _{ i=\frac { N }{ 2 } +1 }^{ N }{ \left( \begin{matrix} N \\ i \end{matrix} \right)  } =\quad \sum _{ i=0 }^{ N }{ \left( \begin{matrix} N \\ i \end{matrix} \right)  } /2,\quad if\quad N\quad is\quad even=2^N/2 = 2^{N-1}$
\end{table*}

The \textit{Simple model} strategy of Table~\ref{tbl:classifier_designs}, represents a classifier design which emphasizes feature reduction in its learning phase. An example of such a classifier is a C4.5 Decision tree \citep{quinlan1993c4} or a Linear SVM with L1-regularization penalty \citep{chang2011libsvm}, which gravitate towards simpler model representations. A representative model in this case, is given by $C:X_i=1$, as the learned classification rule. The probability that a randomly generated $N$-dimensional probe sample will evade $C$ is  0.5, as the prediction space is divided into two halves, on this feature $X_i$. The adversarial certainty (Table~\ref{tbl:classifier_designs}) refers to the confidence that an adversary has about the defender's training data, given that an attack sample is successful in evading $C$. In the case of the simple model, the adversary is fairly uncertain, as the training data sample $L$ could be any one of the $2^{N-1}$ probe samples, which provide successful evasion.

The \textit{Complex models} rely on aggregating feature information, to make the learned models more complex  (i.e., have more coefficients and important features). The one class classifier strategy is a complex model and comes in two flavors: a restrictive boundary around the legitimate training data samples, and a restrictive boundary around the malicious training data samples. In the former case, the model is the hardest to evade, as an adversary will need to simultaneously evade all $N$ features, to gain access to the system. This provides additional security from attack onset, by reducing evasion probability to 1 in $2^N$. However, it also leads to an adversarial certainty of 100\%, implying that any successful attack will lead to a data nullification and subsequent inability of the defender to recover from such attacks. Thus, as opposed to common belief in cybersecurity \citep{onoda2012analysis}, a more restrictive classifiers could be a bad strategy, when considering a dynamic and adversarial environment. The one class model on the set of malicious training data, represents the other end of the spectrum, where a boundary is drawn around the malicious class samples. This model makes evasion easier, but ensures high adversarial uncertainty. This is also undesirable, as evasion can be performed by changing any of the $N$ features. It could however benefit defenders facing multiple disjoint adversaries \citep{kantchelian2013approaches, sculley2011detecting}. 

A popular design strategy in adversarial domains, focuses on integrating multiple orthogonal feature space information to make a complex model, which is robust to changes in a subset of the features \citep{biggio2008adversarial,biggio2010multiple,biggio2010multipleattack}. A feature bagging ensemble, which uses majority voting on its component model's predictions, is a popular choice in this category. This model is resilient to attacks which affect only a few features at any given time. Most robust learning strategies were evaluated against a targeted exploration attack, where the adversary starts with a predetermined attack sample and aims to minimally alter it, so as to evade the classifier $C$. By measuring adversarial cost in terms of the number of features which need evasion, the efficacy of feature bagged models was demonstrated \citep{lowd2005adversarial}, as robust models by design will require a majority of the features to be modified, for a successful evasion.  This reasoning is not valid for the case of an indiscriminate exploratory attack, where an adversary is interested in generating any sample which evades $C$ without any predefined set of attack samples \citep{tsethi2016}. In this setting, the impact of feature bagged ensemble is similar to that of a simple model, as seen from the evasion probability and adversarial certainty computations of Table~\ref{tbl:classifier_designs}. 

An additional design strategy, relies on randomness to provide protection. By training models on multiple different subspace of the data and then randomly choosing one of the models to provide the prediction at any given time, these model aims to mislead attackers, by obscuring the feedback from the black box model $C$. Although this strategy has a higher perceived evasion resistance (Table~\ref{tbl:classifier_designs}), security through obscurity has shown to be ineffective when faced with indiscriminate probing based attacks \citep{vorobeychik2014optimal}.

\subsection{Using Adversarial Margin Density (AMD) to approximate adversarial uncertainty }
\label{sec:amd}

The use of adversarial certainty in Table~\ref{tbl:classifier_designs}, was illustrated by using a random probing attack on a binary feature space. Here, the idea of adversarial uncertainty is expanded and a methodology for its computation and usage is presented, for empirical evaluation. 

Adversarial uncertainty is essentially the uncertainty on the adversary's part, due to the non availability of the original training dataset. For the task of classification, this uncertainty refers to the inability of the adversary to successfully reverse engineer the impact of all the important features, on the prediction task. We measure this uncertainty using the notion of Adversarial Margin Density (AMD), as defined below. The concept of margin density was first introduced in \citep{sethi2015don,sethi2017reliable}, where it was used to measure the uncertainty in data samples, as a method to detect drifts from unlabeled data. It was shown that robust classifiers defined regions of uncertainties (called margins or blindspots), where samples fall as a result of partial feature space disagreement. We extend this notion to the domain of adversarial evaluation, by using it to define adversarial uncertainty. An attacker that impacts only a minimal set of feature, so as to be sufficient to evade the defender's classifier $C$, will lead to a large margin density. This is due to the increased disagreement between orthogonal models, trained from the original training dataset. Conversely, an attacker with a low margin density is indicative of one who was successful in evading a large set of informative features. To this end, the notion of Adversarial Margin Density (AMD) is defined here, to capture adversarial uncertainty on the set of evaded features. 

\newtheorem{definition}{Definition}[section]
\theoremstyle{definition}
\begin{definition}{\textbf{Adversarial Margin Density (AMD):}}
The expected number of \textit{successful} attack samples (attack samples classified by the defender as \textit{Legitimate}), which fall within the defender's margin (i.e., the region of \textit{critical uncertainty}), defined by a robust classifier (one that distributes feature weights) on the training data.
\label{def:amd}
\end{definition}

\begin{figure}[t]
  \centering
  \includegraphics[width=0.95\linewidth]{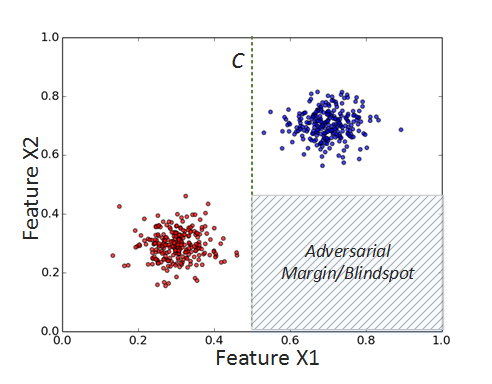}
   \caption{Illustration of adversarial margin. It is given by the region of space which leads to evasion of defender's classifier $C$, but does not lead to evasion of all the informative features.}
  \label{fig:amd}
\end{figure}

The definition of AMD highlights the following two concepts: a) The AMD is computed only on the attack samples which successfully evade the classifier $C$, and b) We measure only the region of critical uncertainty (margin/blindspots), predefined for a classifier trained on the original training data. The AMD measures the uncertainty by quantifying the attack samples which evade the classifier $C$, but do not have high certainty about the entire feature space. This is demonstrated in Figure~\ref{fig:amd}, where the adversarial margin is given by the region of space where the defender $C$ is evaded, but the feature $X2$ is still not successfully reverse engineered by the adversary. This causes the attacks to fall within blindspots of a robust classifier (i.e., one that distributes feature weights). For an attack to have a low uncertainty, it will have to avoid falling in the blindspots, which is possible only if an attacker can successfully evade a significant portion of the feature space (given by critical uncertainty), simultaneously.

The idea behind the AMD is presented in Definition~\ref{def:amd}. An implementation methodology which can be used to compute AMD is presented next. A random subspace ensemble is considered for defining the robust classifier, over the training dataset, as this classifier is general in its design and does not make assumptions about usage of any specific classifier type \citep{ho1998random}. The AMD is computed using Equation~\ref{eqn:amdrs}.

\small
\begin{equation}
\begin{split}
  AMD = \frac { \sum { { S }_{ E }(x) }  }{ \left| x \right|  } ;\quad \forall x\in X_{Attack}:C(x)~ is~Legitimate   
\\
where, \\
{ S }_{ E }(x)=\begin{cases} 1, if\quad \left| { p }_{ E }({ y }_{ Legitimate }|x)-{ p }_{ E }({ y }_{ Malicious }|x) \right|\le { \theta  }_{ Margin } \\ 0, otherwise \end{cases}
\end{split}
\label{eqn:amdrs}
\end{equation}
\normalsize

The AMD is measured only on the attack samples ($X_{Attack}$) which successfully evade the defender's classifier $C$, at test time. These set of samples are evaluated to determine if they fall within the region of critical uncertainty given by the parameter $\theta_{Margin}$. By adjusting the parameter, the sensitivity of measurements and tolerance for adversarial activity, can be tuned. A value of 0.5 is typically considered effective for most scenarios \citep{sethi2017reliable}. Samples falling in this region of high disagreement (given by parameter $\theta_{Margin}$) are considered to be cases where there is critical uncertainty between the constituent models of the ensemble. Since the ensemble is trained as a feature bagged model, this disagreement is due to different feature value distributions, than the benign training samples, caused by adversarial uncertainty. The ensemble $E$ is taken to be a random subspace ensemble. In the experimentation here, we consider a feature bagged ensemble of 50 base Linear-SVM models, each with 50\% of the total features, randomly picked. In Equation~\ref{eqn:amdrs}, $p_E(y|x)$ is obtained via majority voting on the constituent models, and represents disagreement scores for each sample $x$. The AMD is always a ratio between 0 and 1, and a higher value indicates a high adversarial uncertainty. By extension, a higher AMD also indicates an increased ease of unsupervised detection, and subsequent recovery from attacks.

\subsection{Evaluating dynamic effects of evasion by a sophisticated adversary}
\label{sec:high_confidence}

To fully understand the capabilities of an adversary, it is necessary to assume that we are facing a strong adversary, who uses all information and tools at its disposal. To this end, we extend the Anchor Points(AP) attack framework of \citep{tsethi2016}, to simulate a high impact sophisticated adversary, which utilizes its probing ability to generate a high confidence attack on the defender's model. While the framework of \citep{tsethi2016} was developed as a proof of concept, to demonstrate the vulnerability of classifiers to evasion attacks, the proposed attack in this section focuses on the ability of an attacker to not only evade but also cause long term damages, by preventing attacks from being detected or recovered from. Attacks of high adversarial certainty are simulated, which are capable of using probed data to fall outside the margins of the defender. This extension of the AP framework, will allow for testing against a sophisticated adversary, and will enable us to better understand the severity of exploratory attacks possible at test time. Since this extension maintains the original black box assumptions of the defender's model, it provides for analyzing the impact of severe exploratory attacks, which are possible to carry out against a classifier, by an adversary.

\begin{figure}[t]
  \centering
  \includegraphics[width=0.95\linewidth]{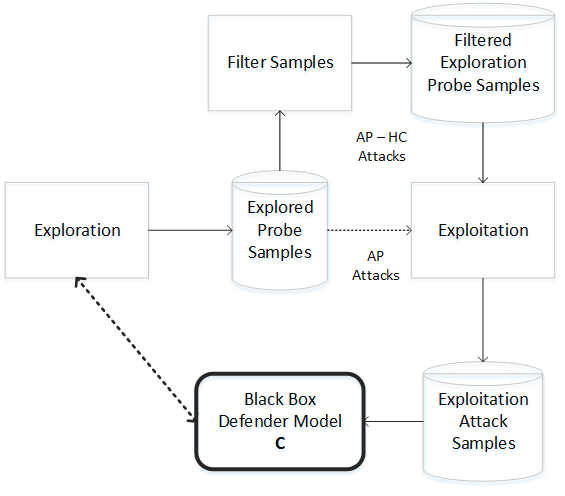}
   \caption{AP based evasion attack framework, with a high confidence filter phase. }
  \label{fig:filter_framework}
\end{figure}

The proposed extension to the AP framework is depicted in Figure~\ref{fig:filter_framework}. Specifically, the filter strategy is developed as an extension to the Anchor Points attacks (AP) of \citep{tsethi2016}. The anchor points exploration samples obtained are first sent to a filter phase, where the low confidence samples are eliminated, before the exploitation phase starts. This filtering leads to a reduced size of exploration samples, for which the adversary has high confidence that they do not fall in the defender's margins. This approach is called the Anchor Points - High Confidence (AP-HC) strategy, and it is developed as a wrapper over the AP framework, making it easy for extension to other data driven attack strategies as well. The AP-HC strategy will simulate adversaries who are capable of utilizing all the probe-able information, to launch attacks which generate low Adversarial Margin Density (AMD), on the defender's part.

\begin{figure*}[t]
  \centering
  \includegraphics[width=0.9\linewidth]{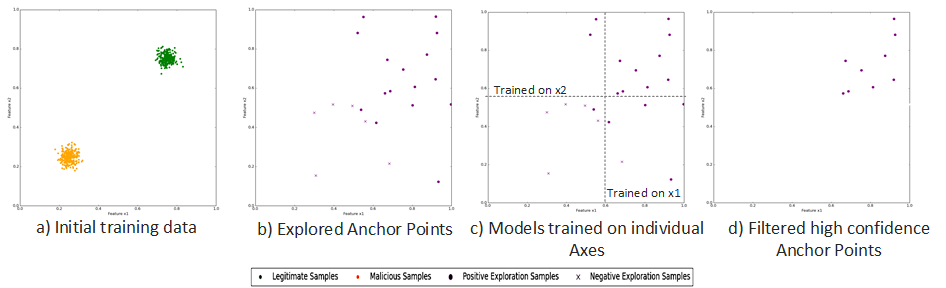}
   \caption{Reducing adversarial uncertainty via filtering, in anchor points attacks. After the exploration phase in b), the adversary trains models on individual feature subspaces (X1 and X2). It then aggregates this information to clean out samples in low confidence areas (C:X1$\ne$ C:X2). The final set of filtered high confidence samples are shown in d).}
  \label{fig:filtered}
\end{figure*}

In the filtering phase of the Figure~\ref{fig:filter_framework}, the adversary  relies on stitching together information made available by the defender's black box, to launch attacks which evade a large number of features simultaneously. The adversary does so by integrating information learned in the exploration phase, about the impact of different subsets of features, on the prediction outcome. It then filters out samples which are good for evasion, but only result in evading a small subset of features. This will result in an attack exploitation, which would provide to high adversarial certainty. The idea is illustrated in Figure~\ref{fig:filtered}, on a 2D synthetic dataset.   In this example, there are two sets of features $X1$ and $X2$, both of which are informative for the classification task, as $X1>0.5 $ or $X2>0.5$ both lead to the high quality defender models. Consider the defender's model to be $'X1>0.5 \vee X2>0.5'$, which leads to an adversarial uncertainty of 1/3, since the training data could be in any of the 3 quadrants where the samples are perceived as \textit{Legitimate}. A naive adversary using the AP framework, can launch an attack using samples where atleast one of the two feature $X1$ or $X2$ is greater than 0.5. However, an adversary using the high confidence filter attack will combine this orthogonal information and launch attack samples only if both conditions are satisfied (i.e., $X1>0.5$ and $X2>0.5$).  In doing so, the adversary is learning the orthogonal information about the prediction landscape and aggregating this information to avoid detection by the defender. 

The high confidence filtering strategy illustrated in Figure~\ref{fig:filtered}, where the initial explored samples for the AP attacks (b), are filtered using the aforementioned information aggregation technique, to generate a high confidence set of exploration samples (c). The adversary does so by using the exploration samples in b), to train orthogonal models, and admit only those samples which have high consensus (low uncertainty), based on the trained model. The filtered exploration samples are then used in the exploitation phase, and as can be seen in Figure~\ref{fig:filtered}, these attacks have an AMD of 0, as none of the attack samples fall inside the margin of the classifier.

In the analysis presented in Table~\ref{tbl:classifier_designs}, it was shown that complex feature bagged ensembles models have a low adversarial certainty ($1/2^{N-1}$, for $N$ dimensional binary feature space). This made it more secure in dynamic environments, when compared to the restrictive one class classifier model, where the adversarial certainty was 1. However, similar to a one class classifier, the robust ensemble model also makes a majority of the information available, to be probed by a patient adversary, as demonstrated in Figure~\ref{fig:filtered}. This information is not directly available (as in the case of a one class classifier, where evasion leads to an adversarial uncertainty of 0), and does not directly lead an adversary to the space of training data. However, the AP-HC filtering step can cause the adversary to stitch together information made available by such ensemble models, to launch a potent attack which could leave the defender helpless. A generic approach to extend the intuition of Figure~\ref{fig:filtered}, to high dimensional spaces, is presented in the following section.

\subsubsection{The Anchor Points - High Confidence (AP-HC) approach}
\label{sec:ap_hc}

The AP-HC attack strategy is presented in Algorithm~\ref{algo:amd}. The algorithm receives the exploration samples from the AP framework, and then uses the filtering approach of Figure~\ref{fig:filtered}, to clear out samples of low confidence. The adversary does so by training a robust classifier, such as a random subspace ensemble, from the probed exploration samples. The trained model is then used to identify samples of low confidence, as these are the ones which have low consensus among the feature bagged ensemble's models. The resulting set of filtered exploration samples $D_{Explore-HC}$, from Algorithm~\ref{algo:amd}, is then used in the exploitation phase of the AP framework,  to launch the attacks. This methodology is implemented completely on the adversary's side, while maintaining the same black box assumption about the defender, as presented by the AP framework in \citep{tsethi2016}. As such, it serves as a methodology for thorough analysis of adversarial capabilities, to better design secure machine learning frameworks. The purpose of the AP-HC attack approach is to highlight the effects of making excessive information available to the adversary, and the resulting space of possible attacks which could be launched by it, using purely data driven methodologies. 

\begin{algorithm}[t]
\SetKwInOut{Input}{Input}
\SetKwInOut{Output}{Output}
 \Input{ Exploration Samples from AP framework - $D_{Explore}$, Filter threshold $\theta_{Adversary\_confidence}$}
 \Output{High confidence exploration samples $D_{Explore-HC}$}

 E $\leftarrow$ Train random subspace samples from $D_{Explore}$
 
 $D_{Explore-HC}$ = $\emptyset$
  
 \For{sample in $D_{Explore}$}{
 	
 	\If {$|p_E(y_{Legitimate}|sample)-p_E(y_{Malicious}|sample)|\ge\quad \theta_{Adversary\_confidence}$ }
 	{
 		$D_{Explore-HC} \quad \cup sample$
 	}}
 \Return $D_{Explore-HC}$
\caption {AP based high confidence filter attacks (AP-HC).}
\label{algo:amd}
\end{algorithm}

The parameter $\theta_{Adversary\_confidence}$, controls the filtering operation in Algorithm~\ref{algo:amd}. Since an adversary is interested only in high confidence attack samples, we consider a high confidence threshold of $\theta_{Adversary\_confidence}$=0.8. A higher confidence threshold will allow for more stringent filtering of samples. This heuristic filtering approach will be used to demonstrate the capabilities of an adversary to evade detection, by gaining more certainty about the location of the training data. This approach will be used to highlight innate vulnerabilities in seemingly secure designs, such as the robust feature bagged ensemble strategy, and will be used for thorough analysis of the impact of classifier design on adversarial capability. 

\section{Experimental evaluation and analysis}
\label{sec:experimentation_5}

In this section, the impact of different classifier design strategies, on adversarial capabilities, is evaluated. Moving beyond accuracy, the effects of attacks from a dynamic-adversarial perspective, is analyzed. The idea of adversarial margin density is considered, to account for detect-ability and retrain-ability from attacks. The adversary is considered to be capable of using machine learning to meet its needs. In Section~\ref{sec:experimental_setup_5}, the protocol and setup of experiments performed in this section, is presented. Effect of using a restrictive one class model for the defender is presented in Section~\ref{sec:oneclass_impact}. Impact of using a robust feature bagged ensemble and that of using randomization based models in presented in Section~\ref{sec:robust_impact} and Section~\ref{sec:random_impact}, respectively. Discussion and analysis is presented in Section~\ref{sec:classifier_discussions}.

\subsection{Experimental setup and protocol}
\label{sec:experimental_setup_5}

The basic \textit{Anchor Points} framework of \citep{tsethi2016}, is considered for generating attacks on the defender's black box model, with the modifications presented in Section~\ref{sec:ap_hc}. The parameters of the AP attacks are taken without any modifications, to ensure consistent analysis and extension of the attack paradigms. The defender is considered to be a black box by the adversary, with no information about its internal working available. The only interaction the attacker has with the defender, is by means of submitting probe samples, and receiving tacit Accept/Reject feedback on them. 

Adversarial margin density proposed in Section~\ref{sec:amd}, is used for evaluating uncertainty of the attacks. A $\theta_{Margin}$=0.5,  for measuring the AMD (as motivated by analysis in \citep{sethi2017reliable}), is considered. A random subspace model with 50 Linear SVMs (regularization constant c=1, and 50\% of the features in each model), is taken for the AMD computation. Also, the effects of filtering by an adversary is evaluated in this section, with a $\theta_{Adversary\_confidence}$=0.8. A random subspace model with 50 Linear SVMs (regularization constant c=10, and 50\% of the features in each model), is chosen for the filtering task. A high regularization constant ensures that the models do not overfit to the exploration samples, as the adversary's goal is not to fit the model to the explored samples, but to learn from it about the region of high confidence. This also makes it more robust to black box feedback noise and stray probes. 

Description of datasets used for evaluation is presented in Table~\ref{tbl:5_datasets}. The synthetic datasets is a 10 dimensional dataset, with two classes. The \textit{Legitimate} class is normally distributed with a $\mu$=0.75 and $\sigma$=0.05, and the \textit{Malicious} class is centered at  $\mu$=0.25 and $\sigma$=0.05, across all 10 dimensions. This datasets provides for controlled experimentation and analysis, by exhibiting 10 informative features. The CAPTCHA dataset is taken from \citep{d2014avatar}, and it represents the task of classifying mouse movement data for humans and bots, for the task of behavioral authentication. The phishing dataset is taken from \citep{Lichman:2013}, and it represents characteristics for malicious and benign web pages. The digits dataset \citep{Lichman:2013} was taken to represent a standard classification task. The multidimensional dataset was reduced to a binary class problem with the class 1 and 7 taken for the Digits17 dataset, and the class 0 and 8 taken for the Digits08 dataset, respectively. In all datasets, the class 0 was considered to represent the \textit{Legitimate} and 1 was taken as the \textit{Malicious} class. For Digits17, the class 7 is considered to be \textit{Legitimate}, and for the Digits08, class 0 is considered \textit{Legitimate}. The data was normalized to the range of [0,1], using min-max normalization, and the features were reduced to a numeric/binary type. The records were shuffled, to eliminate stray effects of concept drifts within them. All experiments are repeated 30 times and average values are reported. The experiments are performed using python and the scikit-learn machine learning library \citep{scikit-learn}.

\begin{table}[t]
\centering
\caption{Description of datasets used for experimentation. }
\label{tbl:5_datasets}
\begin{tabular}{|l|c|c|}
\hline
Dataset & \#Instances & \#Features \\ \hline
Synthetic & 500 & 10 \\ \hline
CAPTCHA & 1886 & 26 \\ \hline
Phishing & 11055 & 46 \\ \hline
Digits08 & 1499 & 16 \\ \hline
Digits17 & 1557 & 16 \\ \hline
\end{tabular}
\end{table}

\subsection{Analyzing effects of using a restrictive one-class classifier for the defender's model}
\label{sec:oneclass_impact}

\begin{figure}[t]
  \centering
  \includegraphics[width=0.75\linewidth]{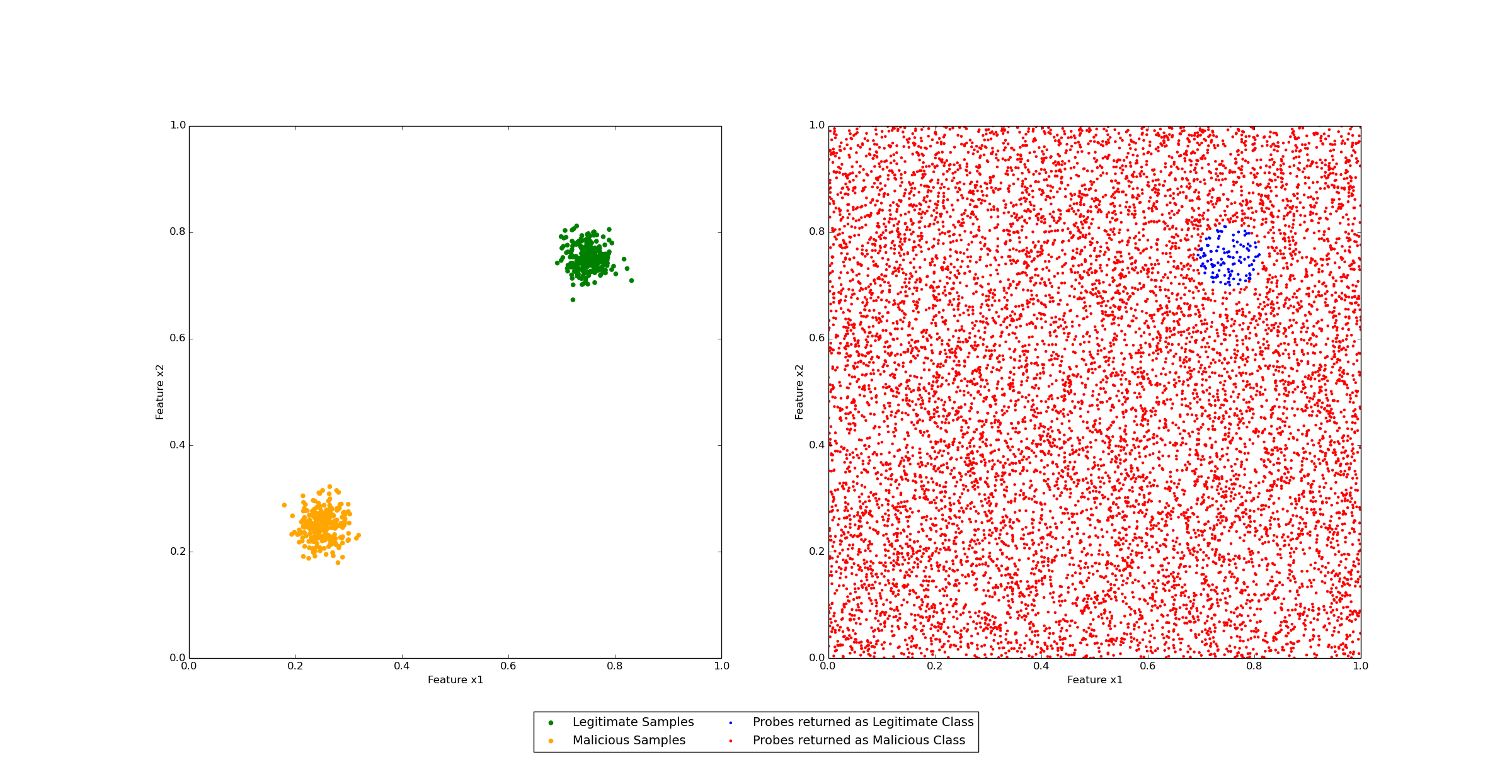}
   \caption{Illustration of prediction landscape of a one-class classifier. Smaller area of the legitimate samples indicate the resilience against probing based attacks. }
  \label{fig:oneclass_space}
\end{figure}

In adversarial domains, restricting the space of samples classified as \textit{Legitimate}, is considered an effective defense strategy \citep{onoda2012analysis, biggio2015one}. By tightly restricting what data points are qualified as legitimate, the ability of random probing based attacks is significantly reduced. This is a consequence of the reduced probe-able feature space area, as shown in Figure~\ref{fig:oneclass_space}, where a one class classifier is trained on the set of legitimate training data points. The feedback obtained from the defender's model, by probing the feature space, is shown. The significantly smaller area of the blue feedback (Figure~\ref{fig:oneclass_space}), is what makes one class classifiers harder to probe and reverse engineer. However, the one class classifier has limitation, which make them unsuitable for an adversarial domain. Firstly, a one class classifier sacrifices generalization ability and it is not always possible to train an accurate model over the available training data \citep{biggio2015one}. Secondly, as seen in our analysis in Table~\ref{tbl:classifier_designs}, a one class classifier could cause high adversarial certainty, making attack detection and recovery difficult. Most works on the security of classifiers advocate inclusion of all orthogonal information, to make the system more restrictive and therefore more secure \citep{onoda2012analysis, biggio2015one,papernot2016towards,rndic2014practical,wang2015robust,biggio2010multiple}. However, these works approach the security of the system from a static perspective. Evaluating a one class classifier provides valuable insights into the inefficacy of using restrictive models, in a dynamic adversarial environment. While recent works present various novel ideas for integrating feature information \citep{biggio2015one,wozniak2014survey}, a one-class classifier serves as an ideal representative approach, in which all information across all features is used in securing the system. 

\begin{figure}[t]
\centering
\subfloat[Use of a one class classifier by the defender causes reduced attack rate, but leads to increased training data corruption and leakage.]{\includegraphics[width=0.95\linewidth]{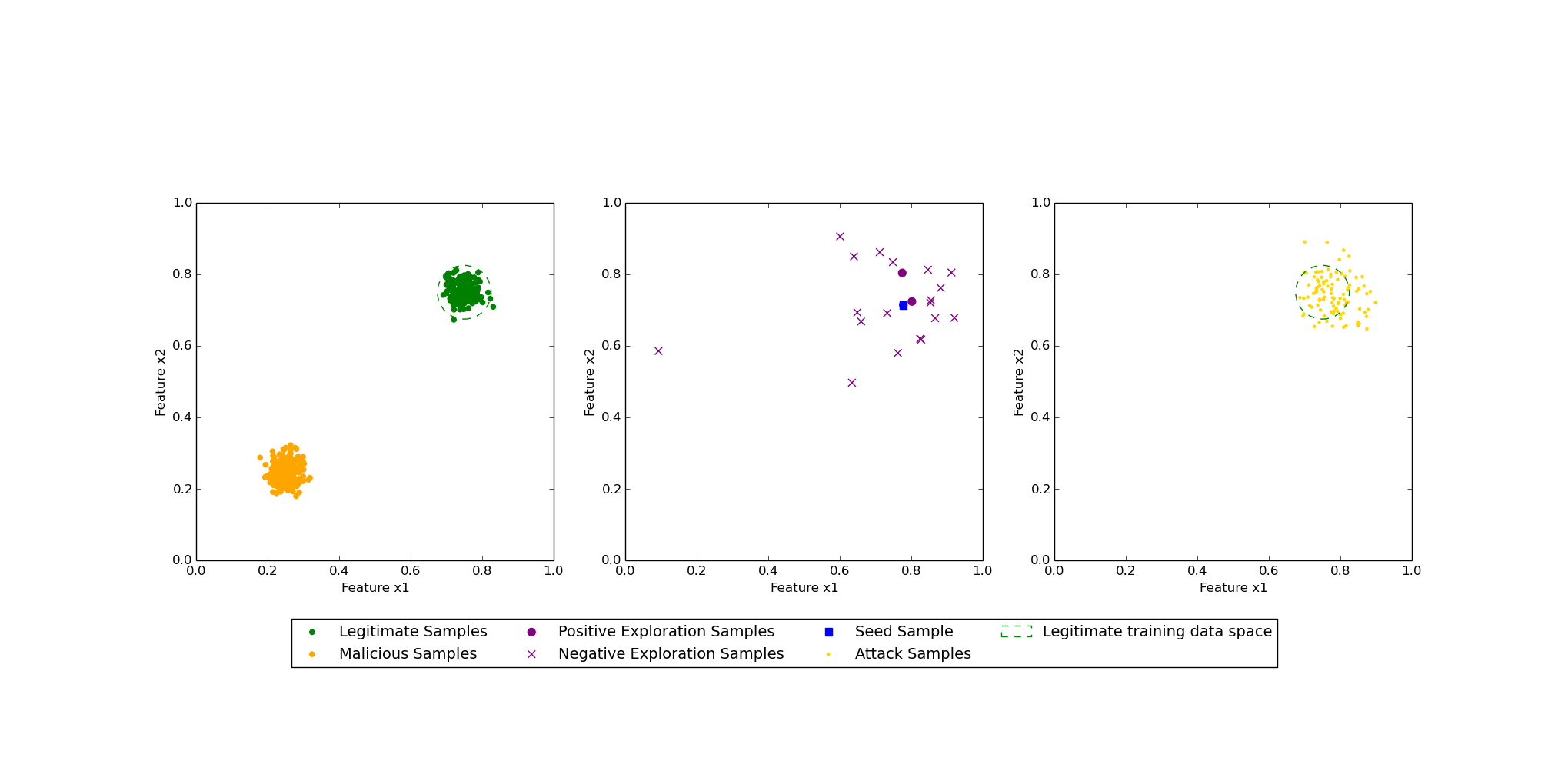}}\\
\subfloat[Two class classifiers ensures less data leakage, but makes evasion easier. ]{\includegraphics[width=0.95\linewidth]{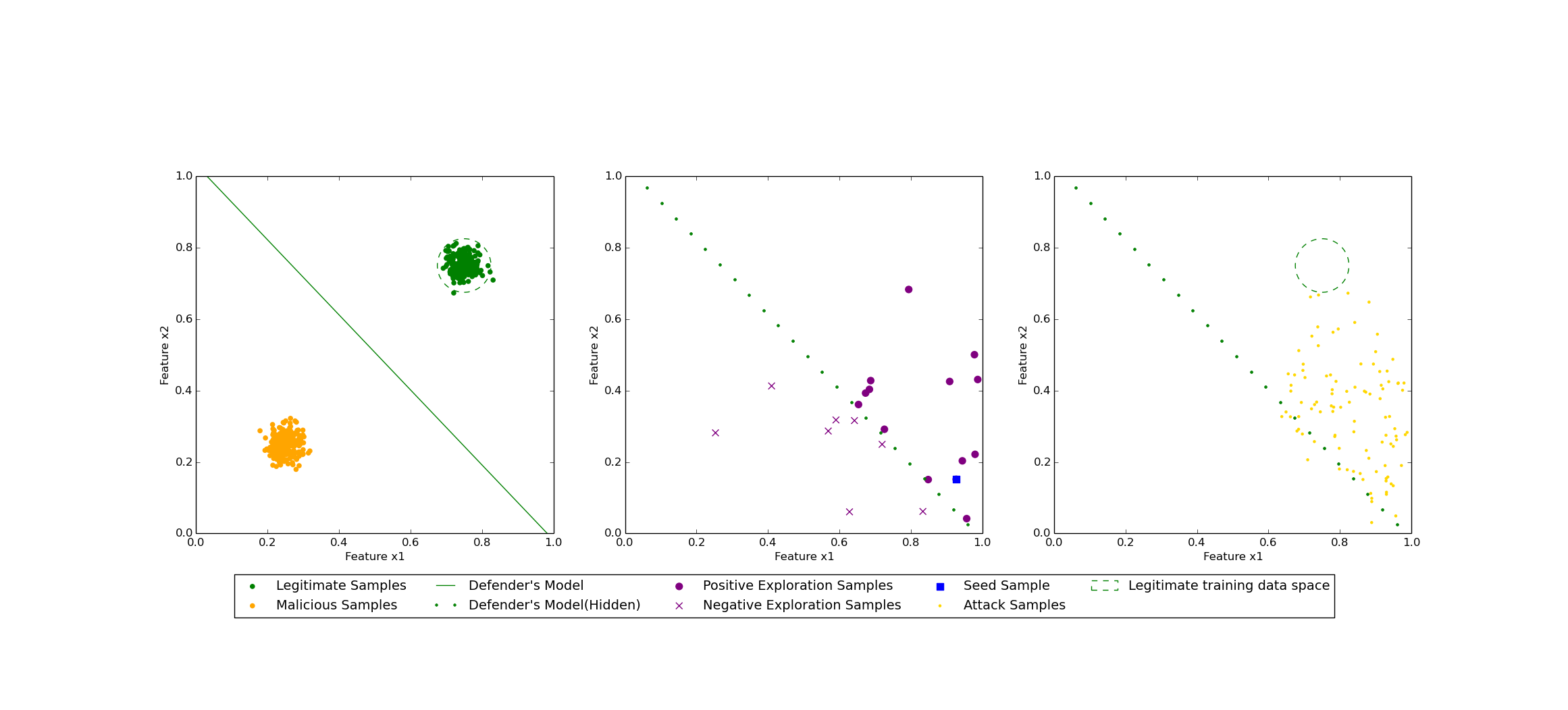}}
\caption{Illustration of AP attacks on a synthetic 2D dataset, using a restrictive one class classifier and a generalized two class classifiers, for the defender's model.}
\label{fig:oneclass_desc}
\end{figure}

In a dynamic environment, it is necessary to maintain adversarial uncertainty, so as to ensure that attacks can be recovered from. A one-class classifier, being overly restrictive, leads an attacker directly to the space of the legitimate training data. Although this classifier design makes the adversary expend greater effort to evade the system, once evaded the attacks are indistinguishable from the benign traffic entering the system. This is illustrated in case of a 2D synthetic data in Figure~\ref{fig:oneclass_desc}.  Here, the anchor points attack is used to generate attacks, on two different classifier designs: a) A one class classifier on the legitimate training data (SVM with parameters: $\nu$=0.1, RBF kernel and $\gamma$=0.1), and b) A two class linear classifier model (Linear SVM with L2-regularization, c=1). The attack used 20 samples for exploration ($B_{Explore}$) and generated 40 attack samples, in each case \citep{sethi2017data}. It is seen in Figure~\ref{fig:oneclass_desc}, that the attack leads to a large number of samples occupying the same region as that of the legitimate samples, in case of the restrictive one class defender model (Figure~\ref{fig:oneclass_desc}a)). This will cause problems in a dynamic environment,  as retraining to thwart attacks is close to impossible in this case. Also, detecting such attacks is difficult, due to the increased similarity with benign input. In case of the two class model (Figure~\ref{fig:oneclass_desc}b)), a larger data space is perceived as legitimate, due to the generalization provided by these models, leading to attacks which are farther from the training data space. This illustrates the ability of classifier model designs, to influence the severity of attacks on the system, and to cause higher adversarial uncertainty.  

\begin{table*}[t]
\centering
\caption{Results of experimentation on one-class and two-class defender models. Training accuracy, Effective Attack Rate (EAR), Data Leakage (DL) and Adversarial Margin Density (AMD), are presented for comparison. }
\label{tbl:oneclass}
\scalebox{0.9}{
\begin{tabular}{|c|c|c|c|c|c|}
\hline
Dataset & Defender's Model & \begin{tabular}[c]{@{}c@{}}Training \\ Accuracy\end{tabular} & EAR & Data Leak & \begin{tabular}[c]{@{}c@{}}Adversarial Margin \\ Density (AMD)\end{tabular} \\ \hline
\multirow{2}{*}{\begin{tabular}[c]{@{}c@{}}Synthetic\\ (2 features)\end{tabular}} & Restrictive one class & 97.4 & 0.56 & 0.67 & 0 \\ \cline{2-6} 
 & Robust two class & 100 & 0.89 & 0 & 0.85 \\ \hline
\multirow{2}{*}{\begin{tabular}[c]{@{}c@{}}Synthetic \\ (10 features)\end{tabular}} & Restrictive one class & 97.6 & 0.04 & 0.67 & 0 \\ \cline{2-6} 
 & Robust two class & 100 & 0.99 & 0.1 & 0.28 \\ \hline
\multirow{2}{*}{CAPTCHA} & Restrictive one class & 96.4 & 0.88 & 0.23 & 0 \\ \cline{2-6} 
 & Robust two class & 100 & 0.99 & 0.01 & 0.14 \\ \hline
\end{tabular}}
\end{table*}

Experimental evaluation of the synthetic 2D dataset, and two additional datasets from Table~\ref{tbl:5_datasets}, is presented in Table~\ref{tbl:oneclass}. Here, the metric of \textit{Effective Attack Rate} (EAR) is taken to measure the vulnerability of the the defender's model, to anchor points attacks. EAR measures the percentage of attack samples which are incorrectly classified by the defender's classifier model \citep{tsethi2016}. Additionally the following two metrics are introduced, to measure the effectiveness of attacks in a dynamic environment - \textit{Data Leakage} and the \textit{Adversarial Margin Density} (AMD). Adversarial margin density was introduced in Section~\ref{sec:amd}, as a measure for approximating adversarial uncertainty over the defender's training data. Data leakage is introduced here as an adhoc metric for evaluating the loss of private data, in a one class classification setting. Data leakage is measured by developing a one class classifier on the space of legitimate training samples, and then measuring the number of attack samples which are incorrectly classified by this classifier. Data leak is used to measure the proximity of the attack samples, to the original space of legitimate training samples, with a large value indicating the ability of the adversary to closely mimic the legitimate samples.  A one class SVM model (parameters: $\nu$=0.1, RBF kernel, $\gamma$=0.1), is used to measure the data leakage metric. The metrics were computed for the case of the one class and the two class defender's model, as shown in Table~\ref{tbl:oneclass}.

The Effective Attack Rate (EAR), is significantly lower for the one class classifier ($\Delta$=0.46 on average, from Table~\ref{tbl:oneclass}). This is a result of the stricter criterion for inclusion into the legitimate space, as imposed by the complex and restrictive classifier boundary. However, this comes at the cost of an increased possibility of data leakage and lower adversarial uncertainty. From Table~\ref{tbl:oneclass}, it can be seen that the data leak increases sharply for the one class classifier ($\Delta$=0.49, on average), as the restrictive nature of the classifiers leads the adversary to the training data samples.  This causes problems with retraining, loss of privacy, loss of clean training data, and issues with unsupervised attack detection. The data leak metric provides a heuristic way to measure the severity of data nullification attacks. However, the data leak metric is difficult to compute for high dimensional datasets, due to the inability to train an accurate one class classifier, which is needed to measure the data leakage. For this reason the other datasets of Table~\ref{tbl:5_datasets}, are not used for the analysis in this section. Also, these datasets provided low training data accuracy when using a one class classifier, making it unsuitable as a choice for the defender's model. Going forward, the adversarial margin density metric will be used, to indicate the strategic advantage of the defender over the adversary, in detecting attacks and relearning from them. 

The adversarial margin density (AMD) is more general in its applicability, when compared to the data leak measure, and provides a way to indirectly measure adversarial uncertainty and the severity of attacks in dynamic environments. From Table~\ref{tbl:oneclass}, it is seen that the AMD is 0 for all 3 datasets, when using a one-class defender model. This is due to the increased confidence in the location of the training data, once the restrictive model is evaded. Although the one class classifier is secure, as it ensures a low EAR, it leaves the defender helpless once an attack starts. As such, designing for a dynamic environment requires forethought on the defender's part. The experimentation in this section was  presented for illustrating the ill effects of using an overly restrictive model for the defender, and the need to reevaluate the notion of security which relies on generating complex learning models.

\subsection{Analyzing effects of using a robust feature-bagged ensemble for the defender's model}
\label{sec:robust_impact}

\begin{figure}[t]
  \centering
  \includegraphics[width=0.95\linewidth]{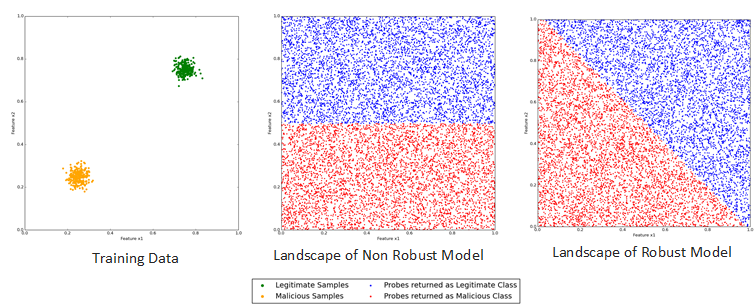}
   \caption{Illustration of prediction landscape using simple and robust models, on 2D synthetic data. \textit{Left}- Initial training data, of the defender. \textit{Center}- L1-regularized linear SVM model for the defender (Non robust). \textit{Right}- L2-regularized linear SVM model for the defender (robust).}
  \label{fig:robust_space}
\end{figure}

Robust models, which advocate complex models involving a large number of informative features, are considered to be effective in safeguarding against targeted exploratory attacks \citep{papernot2016towards,rndic2014practical,wang2015robust,biggio2010multiple,biggio2010multipleattack, wozniak2014survey, stevens2013hardness}. In these attacks, the adversary starts with a predefined set of attack samples, and intends to minimally modify it, so as to evade the defender. Robust models require that a majority of the feature values be mimicked by the adversary, increasing the cost and effort needed to carry out attacks. However, in the case of an indiscriminate exploratory attacks \citep{tsethi2016}, this same line of reasoning is not valid, as an attacker is interested in any sample that causes evasion. In these attacks, the ease of evasion is given by the effective space of prediction, which is recognized as \textit{Legitimate} by the defender. This is because an adversary launching indiscriminate attacks does so by probing the black box model to find samples which would be classified as legitimate. A robust model, such as a Linear SVM with L2-regularization, incorporates information conveyed by a majority of the features, from the training dataset, into the learned model. It does so to make a model robust to stray changes, and also to generate wide generalization margins, for better test time predictive performance. In an adversarial environment, the effective area of legitimate samples conveyed by a robust model, is similar to that of a simple model (which aims at reducing the number of features in the model), as shown in Figure~\ref{fig:robust_space}. In Figure~\ref{fig:robust_space}, a non robust linear SVM (with L1-regularization) is considered alongside a robust linear SVM (with L2-regularization). The effective area of legitimate samples (blue) is the same in both cases. This makes the two strategies equivalent in terms of securing against indiscriminate probing based evasion attacks.

The equivalence of robust and non robust models, to indiscriminate probing attacks, is further analyzed by evaluating on 5 datasets in Table~\ref{tbl:robust_results}. The Anchor Points attacks (AP), are performed on two sets of classifiers: a) A robust classifier - Random subspace ensemble with 50 linear SVMs (L1-regularized, each with 50\% of the features randomly selected), and b) A non robust classifier- Single linear SVM (with L1-regularization). The effect of these strategies on the adversarial outcome is presented in Table~\ref{tbl:robust_results}. The Effective Attack Rate (EAR) is seen to be similar for the two cases ($\Delta$=0.01, on average). This demonstrates the equivalence in effects of the two strategies, when it comes to securing against probing based attacks. 

Based on the analysis on a binary feature space in Table~\ref{tbl:classifier_designs}, it was deduced that simple models behave similar to robust models, when attacks are of an indiscriminate nature, as both result in the same adversarial certainty of $1/2^{N-1}$. From the same analysis, it was  deduced that robust models are better than one-class classifiers, in maintaining adversarial uncertainty. The intuition behind this was that the increased generalization of robust models will create uncertainty regarding the exact location of the training data, and the impact of various feature subspaces on the prediction task. However, this intuition relied on the assumption of a naive adversary, whose primary aim is to evade the system only. The proposed high confidence filtering attack strategy (AP-HC) of Section~\ref{sec:high_confidence}, provides a way to simulate a sophisticated adversary, who is capable of utilizing all the probe-able information to launch attacks of high certainty. The result of using this attack strategy on the robust classifier model is presented in Table~\ref{tbl:robust_results}. It is seen that the AMD for the robust classifier significantly reduces, when faced with the high confidence attack. In case of the CAPTCHA dataset and the Synthetic dataset, the adversarial activity will go totally unnoticed, while in case of other datasets, a significant drop in the AMD is observed ($\Delta$=0.38, on average over all datasets). An important observation comes from the fact that the filtering operation was performed completely on the adversary's side, without any help or information from the defender's model. The adversary starts off with the goal of admitting only the most confident exploration samples, and in doing so, it makes it difficult for the defender to detect or stop it. 

\begin{table}[t]
\centering
\caption{Results of Anchor Points (AP) attacks and Anchor Points  -- High Confidence (AP-HC) attacks, on the Effective Attack Rate (EAR) and the Adversarial Margin Density(AMD).}
\label{tbl:robust_results}
\scalebox{0.8}{
\begin{tabular}{|l|c|c|c|c|c|}
\hline
\multicolumn{1}{|c|}{\multirow{2}{*}{Dataset}} & \multirow{2}{*}{\begin{tabular}[c]{@{}c@{}}Training \\ Accuracy\end{tabular}} & \multicolumn{2}{c|}{EAR} & \multicolumn{2}{c|}{AMD} \\ \cline{3-6} 
\multicolumn{1}{|c|}{} &  & \begin{tabular}[c]{@{}c@{}}AP \\ Attacks\end{tabular} & \begin{tabular}[c]{@{}c@{}}AP-HC \\ Attacks\end{tabular} & \begin{tabular}[c]{@{}c@{}}AP \\ Attacks\end{tabular} & \begin{tabular}[c]{@{}c@{}}AP-HC \\ Attacks\end{tabular} \\ \hline
Synthetic & 100 & 0.98 & 0.999 & 0.28 & 0.01 \\ \hline
CAPTCHA & 100 & 0.996 & 0.999 & 0.19 & 0.002 \\ \hline
Phishing & 93.1 & 0.978 & 0.999 & 0.62 & 0.19 \\ \hline
Digits08 & 97.1 & 0.928 & 0.999 & 0.73 & 0.22 \\ \hline
Digits17 & 99.5 & 0.965 & 0.999 & 0.66 & 0.16 \\ \hline
\end{tabular}}
\end{table}

The effectiveness of the high confidence attacks, is due to the availability of all information, to be probed by a sophisticated adversary. The majority voting scheme of the robust ensemble, is reverse engineered by stitching together orthogonal subsets of feature information, to generate attacks which closely mimic the legitimate samples. As such, the robust ensemble methodologies behave similar to one-class classifiers, by being vulnerable to low adversarial uncertainty. Both design approaches rely on incorporating maximum extracted information from the training data, thereby conveying excessive information to an adversary and equipping it with a thorough understanding of the impact of different features to the classification task.

\subsection{Analyzing effects of using randomization in the defender's model}
\label{sec:random_impact}

Robustness via complex learning methodologies aims at increasing the adversary's effort, by making it reverse engineer a large set of features, as seen in the case of one-class classifiers and the feature bagged ensemble. The other advocated approach to dealing with adversarial activity relies on obfuscation via randomization. By randomizing the feedback presented to the attacker, these methodologies aim to mislead adversarial learning from probes, leading to less effective attacks. Randomness can be introduced into multiple classifier systems, particularly the ones using feature bagging. This is done by training multiple models on subsets of features and then using any one of these models to provide prediction at any given time \citep{wozniak2014survey,biggio2010multiple,colbaugh2012predictive}. The idea relies on confusing the adversary by constantly presenting it with a moving target.   In this section, the impact of randomization is analyzed, when used with a feature bagged ensemble, against indiscriminate evasion attacks. 

\begin{figure}[t]
  \centering
  \includegraphics[width=0.95\linewidth]{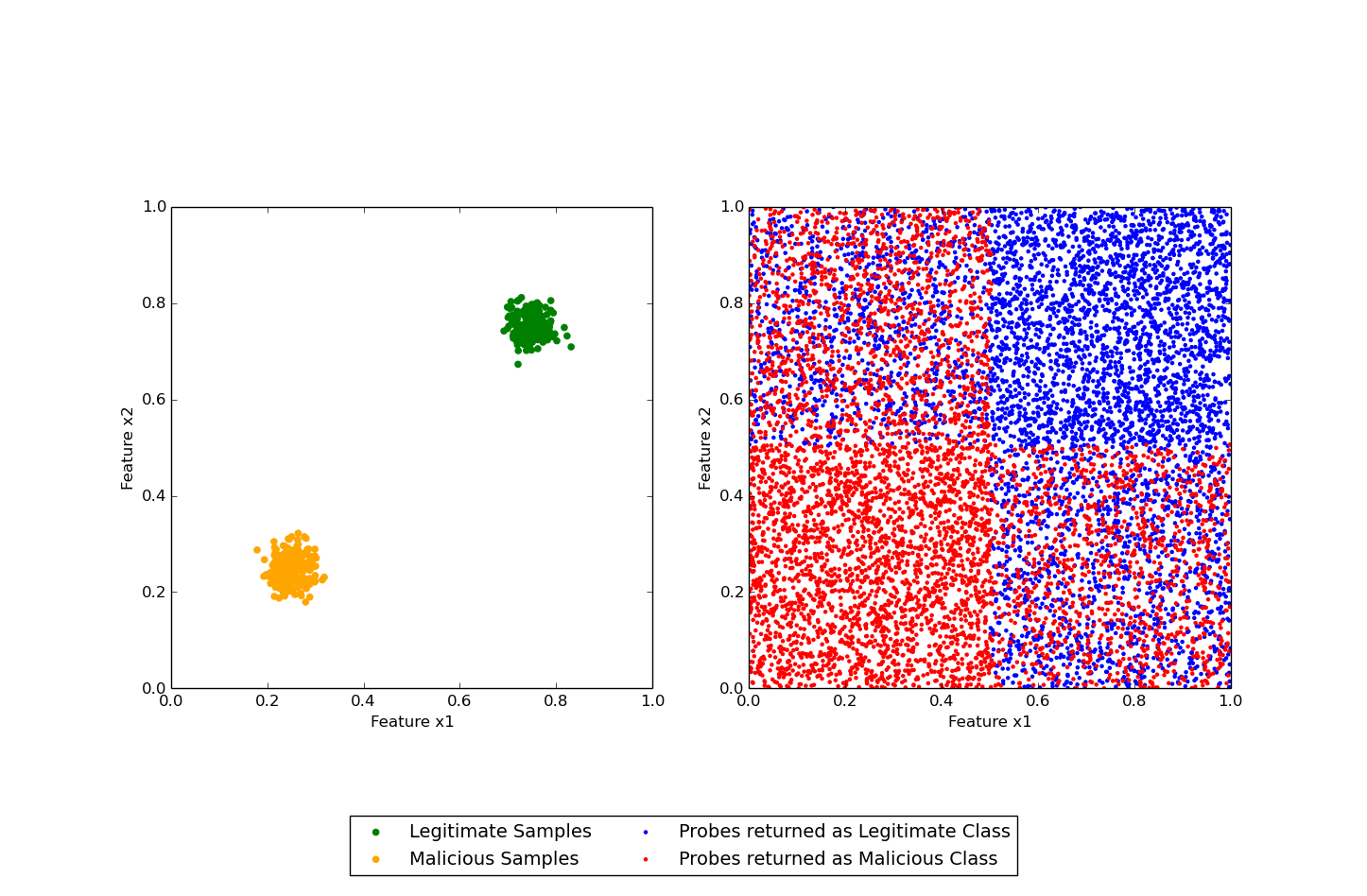}
   \caption{Prediction landscape for the randomized feature bagging ensemble. Blindspots are perceived to be obscured, while high confidence spaces remain consistent across repeated probing.}
  \label{fig:random_space}
\end{figure}

Randomization is introduced into the defender's model, by extending the random subspace ensemble of Section~\ref{sec:robust_impact}, to generate feedback based on the posterior probability for a given sample $X$. The use of random subspace ensemble allows for the randomness be caused due to disagreement between orthogonal feature information. Given a sample $X$, the defender computes the confidence on it, based on majority voting of its component models. This confidence is then used as a probability of prediction, to generate the class label for $X$. As an example, consider a sample $X$ for which the classifier $C$ predicts with 0.8 probability to be in the \textit{Legitimate} class. A standard threshold of 0.5, based on majority voting, will cause the feedback on $X$ to be of class \textit{Legitimate}. To introduce randomness, the defender's model will instead sample number in the range of [0,1] and return feedback \textit{Legitimate}, only if the random number is $>$0.8. As such, randomness is introduced into the classification scheme, while still maintaining the essential predictive properties of the classifier. Here, the adversary may not be aware of the internal scoring or the randomness of the black box model, as it still only experiences the defender as a black box model providing \textit{Legitimate/Malicious} feedback on the submitted probe $X$. The perceived space of the adversary is  shown in Figure~\ref{fig:random_space}. The regions of uncertainty is heavily influenced by randomness, due to the disagreement between models trained on the two features.

\begin{figure}[t]
\centering
\subfloat[Naive adversary]{\includegraphics[width=0.95\linewidth]{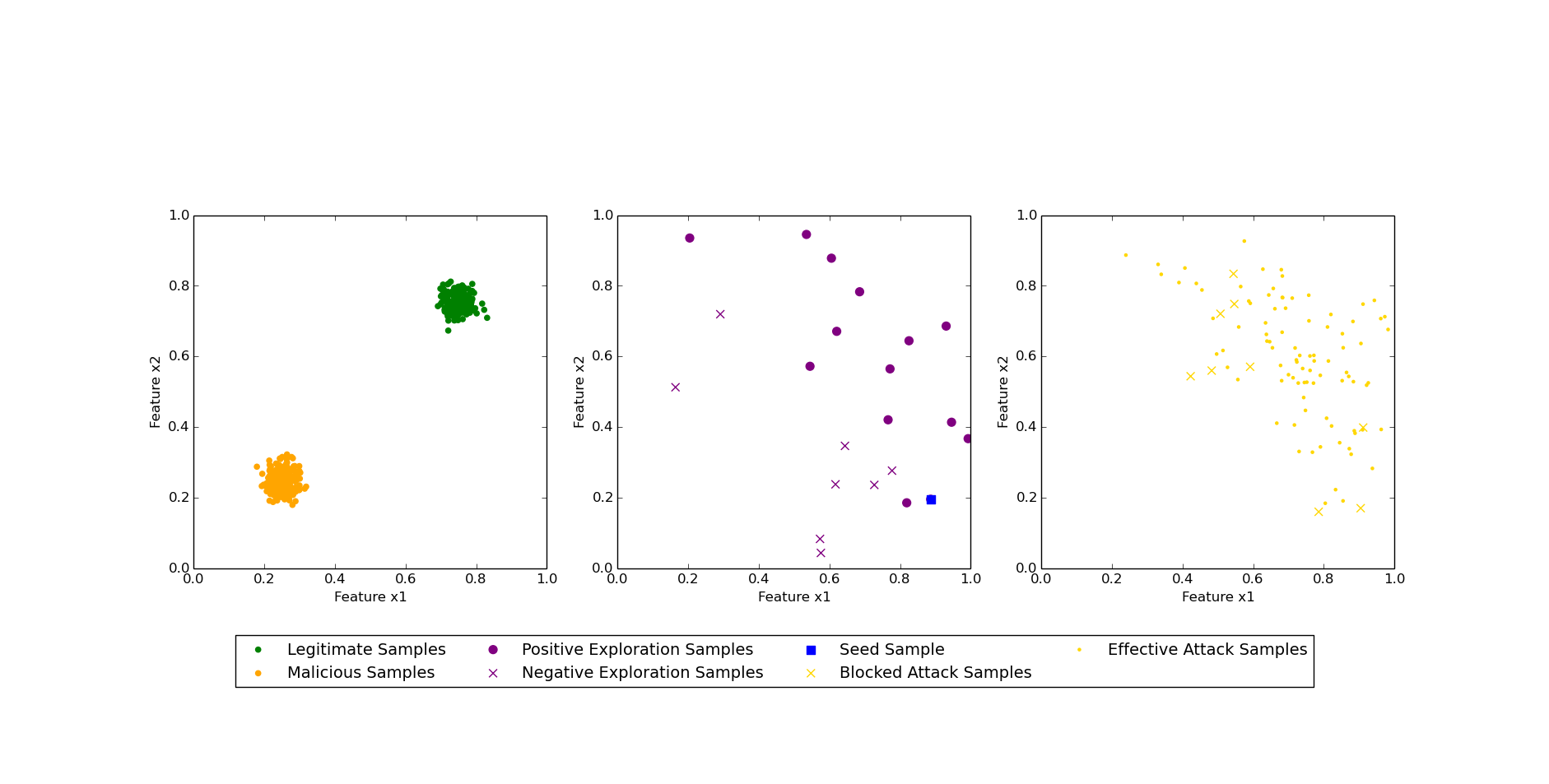}}\\
\subfloat[Randomization aware adversary]{ \includegraphics[width=0.95\linewidth]{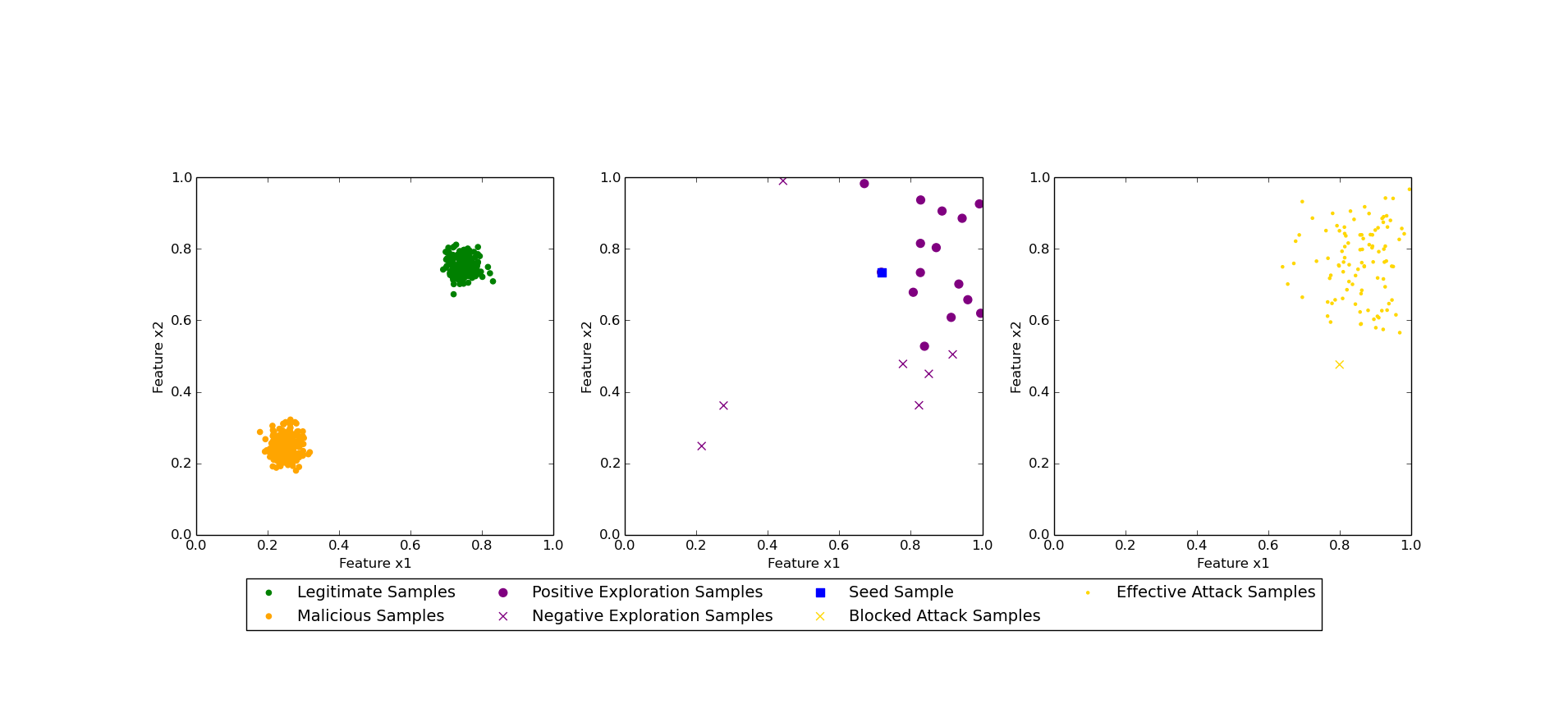}}
\caption{Anchor Points attacks against a randomized defender. \textit{Top-} Naive adversary disregards randomness. \textit{Bottom-} Adversary with confidence filtering, repeated probing of exploration samples is used to weed out samples with inconsistent feedback.}
\label{fig:random_nofilter}
\end{figure}

\begin{figure}[t]
  \centering
  \includegraphics[width=0.95\linewidth]{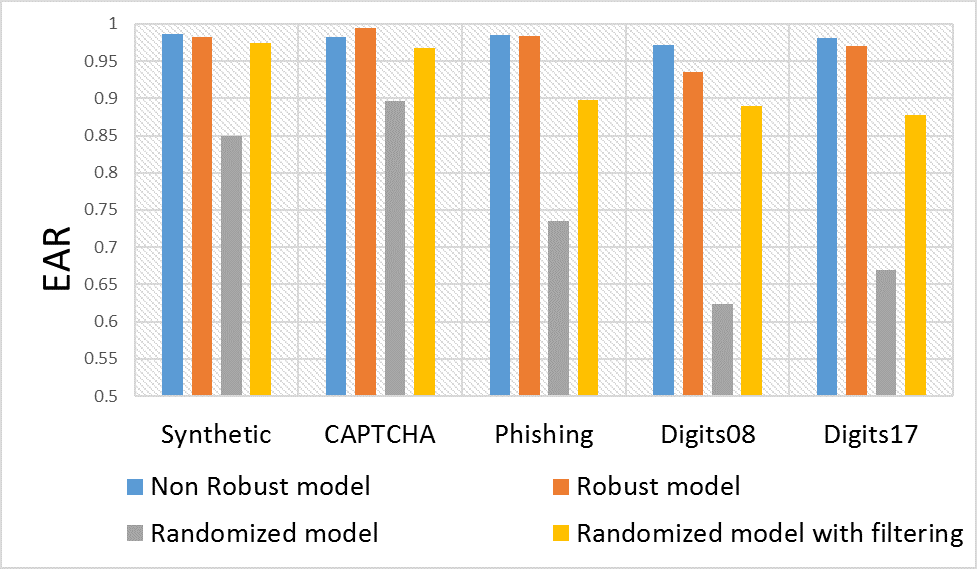}
   \caption{Comparison of Effective Attack Rate (EAR) for  Non robust models, Robust models, Randomization based models, and Randomization models with adversaries capable of filtering low confidence samples. }
  \label{fig:comparison_random}
\end{figure}

The effects of randomness is demonstrated over a synthetic 2D dataset in Figure~\ref{fig:random_nofilter}a), where the anchor points (AP) attacks is used. The misleading feedback from the defender, causes the exploration phase to be corrupted, due to the naive assumption on the adversary's part about the veracity of the defender's feedback. This is presented in Figure~\ref{fig:comparison_random} for the 5 datasets. An average drop of 22.9\% in the Effective Attack Rate (EAR), is seen. 

The analysis of a naive adversary assumes that the defender always returns the correct feedback on the submitted probes. This was seen to result in the attacker being mislead, as seen for the EAR of the naive attacker in Figure~\ref{fig:random_nofilter}b). However, an adversary can become aware of the randomness, by submitting the same probe multiple times and observing different responses on it. Such an adversary can account for randomness in designing its attacks. A simple heuristic strategy is simulated here, to understand the behavior of such an adversary. In the exploration phase, the adversary makes repeated submission on every exploration point, to understand the defender's confidence in the sample. A sample which is closer to the training data, will generate feedback with more consistency, while samples falling in blindspots will be more random in their feedback (Figure~\ref{fig:random_space}). Using this intuition, the the anchor points (AP) attack strategy is modified, to account for smart adversaries capable of dealing with randomness. In the exploration phase, the adversary submits each sample $N_{Retries}$ (taken as 5 in experiments here) times, and accepts the probe to belong to the \textit{Legitimate} class, only if it returns the same feedback for all $N_{Retries}$ times. All other samples are assumed to belong to the $Malicious$ class. By cleaning samples of low confidence, the blindspot exploration samples are removed, making the exploitation phase more potent, as demonstrated in Figure~\ref{fig:random_nofilter}b).

Applying the filtering step allows for simulating an adversary capable of dealing with randomness. The results of such an adversary is shown in Figure~\ref{fig:comparison_random}, where an increase in 16.7\% in the EAR is seen, for such attackers. After this filtering, the attacks are similar to that on a robust model, with only a 5.1\% difference on average. This demonstrates the ineffectiveness of the randomization approaches in providing security, against probing based attacks. There is increased onus on an adversary to use more probes to validate the exploration samples, but if a possible adversary makes this additional investment, incentivized by a high EAR, the randomization approach fails.

\subsection{Why informative models are ill suited for adversarial environments?}
\label{sec:classifier_discussions}

The evaluation of classifier design from the perspective of adversarial uncertainty provides new insights into their vulnerabilities, when applied in a dynamic adversarial environment.  The model design strategies of one-class classifiers, robust feature bagging models and randomized feature bagging models,  are all susceptible to attacks which can leave detection and retraining, intractable. All these methodologies try to include the maximum information from the training data, into the learned model, in order to maintain an information advantage over the adversary. This is seen to be the general trend in the adversarial machine learning research community \citep{papernot2016towards,biggio2010multiple,biggio2014pattern}, where including more information into the models is believed to make it more secure. However, incorporating more information into the learned model means that the adversary will be able to reverse engineer and learn more of this information from the deployed model, via probing. In case of a one-class classifier, which advocates using maximal information to develop a tight boundary around the training data, it was seen that an attacker is led straight to the space of the legitimate data (Section~\ref{sec:oneclass_impact}). In case of Robustness and Randomization, information available is made harder to mimic/extract. In robustness, more features need to be mimicked to gain access, while randomization aims to mislead the adversary's learning. It is seen that both fail against a determined adversary, who given time/resources can learn and leverage all the available information. 

Robust models are also seen to fail at detection and recovery, when faced with an adversary who is capable of generating high confidence attacks. These also stems from the increased availability of probe-able information, to be used by an adversary. These defense strategies relies of unrealistic assumption on the adversary's part, to ensure safety. It is generally assumed that if an attack is too expensive (i.e., requires many probes or the modification of many features), the adversary will give up. While this assumption is the basis of security against targeted evasion attacks, it does not hold in case of indiscriminate exploratory attacks. This section aimed to highlight some of the issues with incorporating excessive information into the defender's model. In an adversarial domain, the attack is a function of the deployed model. A highly informative model will lead to a highly confident attack. It is necessary to reevaluate this central idea of overly informative defender models, and analyze the effects of causing an information gap between the defender and the attacker, for improving security. 

\section{Ensuring long term dynamic security by hiding feature importance}
\label{sec:hidden}

The approaches of \textit{Complex learning} and \textit{Randomization}, rely on including maximum information from the training dataset, into the learned model. As such, they are vulnerable to information leakage via adversarial probing and reverse engineering. Experimentation in Section~\ref{sec:experimentation_5}, demonstrated that a classifier which exposes excessive information about the feature importance is susceptible to high confidence attacks. The increased confidence in attacks leads to issues with adversarial detect-ability, data leakage and retrain-ability. As such, the impact and ability to hide the importance of some of the features, to ensure that the model is shielded from total reverse engineering, is evaluated here. This approach presents a counter-intuitive idea, as it advocates reducing the amount of information incorporated into the deployed model. 

\subsection{Motivation for feature importance hiding}
\label{sec:motivation_hidden}

The notion behind feature importance hiding relies on eliminating a few of the important features from the classification training phase, so as to intentionally misrepresent their importance to an external adversary. By eliminating these features from the classification model, the adversary is made to believe that they are not important for the prediction task. No amount of probing will help the adversary to ascertain the importance of the hidden features, as their importance and informativeness is shielded from the classification task. Although, this does not provide any direct benefits against the onset of evasion attacks, they help in maintaining adversarial uncertainty. An attacker will not be able to completely reverse engineer the importance of all features, no matter the resources/time used by it, as they are not available in the black box model. 

\begin{figure}[t]
  \centering
  \includegraphics[width=0.95\linewidth]{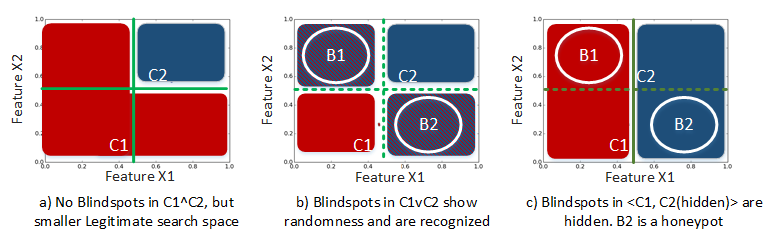}
   \caption{Prediction landscapes as perceived by probing on the defender models. \textit{a)}: Defender model is given as $C1\wedge C2$. \textit{b)}: Defender model given by randomly selecting $C1\vee C2$, to perform prediction.\textit{ c)}: Defender model is given by C1 (trained on feature X1), while C2 is kept hidden to detect adversarial activity. Blindspot(Margin) B2 denotes region for adversarial uncertainty.  }
  \label{fig:honeypot}
\end{figure}

The idea of feature importance hiding, is illustrated in Figure~\ref{fig:honeypot}. The prediction landscapes of the following classifiers are depicted: a) A restrictive classifier aggregating all feature information ($X1\wedge X2$), b) A randomization based classifier which picks its output based on either feature's prediction ($X1\vee X2$), and c) The hidden feature classifier where feature $X2$'s influence is hidden. The choice of these designs affects the adversarial uncertainty, at test time. Based on experimentation in Section~\ref{sec:oneclass_impact} and Section~\ref{sec:random_impact}, we see that restrictive models (a)) and randomization models (b)) are ineffective, as an intelligent adversary can reverse engineer them to generate high confidence attacks.  The robust model of Figure~\ref{fig:honeypot} a) reduces the effective space of legitimate samples, but in doing so it leaks information about the importance of features  $X1$ and $X2$. The randomized model of Figure~\ref{fig:honeypot} b), aims to mislead the adversarial probing, but is vulnerable to repeated probing attacks, as demonstrated in Section~\ref{sec:random_impact}. In case of the hidden feature importance design of Figure~\ref{fig:honeypot} c), an adversary sees a misrepresented view of the prediction landscape. The adversary is led to believe that only feature $X1$ is important to the defender's model. Since $C2$ is kept hidden, the attacker has information about only half the feature space and no amount of probing will help it understand the importance of feature $X2$ to the prediction problem. Attacks generated at test time, will fall under the blue region to make them effective, and when they do so, they have 50\% chance of falling under the blindspot $B2$. $B2$ serves as a honeypot in the learned model, which helps capture attacks.  An important distinction between the robustness and randomization approaches, in comparison to the hidden classifier approach, is that in the former case the defender expects the adversary to be dissuaded by the increased cost of evasion (in terms of probing budget), while in the latter case, the defender is making it unfeasible for an attacker to probe and obtain information necessary to generate a high confidence attack. 

It should be noted that, the idea of hiding feature importance is not in direct violation of the Kerckhoffs's security principle \citep{kerckhoffs1883cryptographie,mrdovic2008kerckhoffs}, which states that security should not rely solely on the unavailability of information, on the part of the adversary. This is because the defender is not using hidden features, but only misleading the adversary into believing that a few of the features are not important to the prediction task. As in Figure~\ref{fig:honeypot}, the attacker is aware of features $X1$ and $X2$, but infers that only $X1$ is important to the classification task. All features are included in the classification model, but by intentionally hiding the importance of a subset of the features, the adversary is misled into generating attacks of partial confidence only (i.e., partially mimics the training legitimate data). 

\begin{table*}[t]
\centering
\caption{Training Accuracy, Effective attack rate (EAR) and Adversarial Margin Density (AMD) under AP-HC attacks, for defender using all features and one which uses only half of the features. }
\label{tbl:hidden}
\scalebox{0.9}{
\begin{tabular}{|c|c|c|c|c|c|c|}
\hline
\multirow{2}{*}{Dataset} & \multicolumn{2}{c|}{\begin{tabular}[c]{@{}c@{}}Training \\ Accuracy\end{tabular}} & \multicolumn{2}{c|}{EAR} & \multicolumn{2}{c|}{AMD} \\ \cline{2-7} 
 & All Features & Hidden Features & All Features & Hidden Features & All Features & Hidden Features \\ \hline
Synthetic & 100 & 100 & 1 & 0.99 & 0.01 & 0.172 \\ \hline
CAPTCHA & 100 & 100 & 0.99 & 0.99 & 0.002 & 0.075 \\ \hline
Phishing & 93.1 & 89.4 & 0.99 & 0.99 & 0.19 & 0.562 \\ \hline
Digits08 & 97.1 & 95.1 & 0.99 & 0.99 & 0.22 & 0.648 \\ \hline
Digits17 & 99.5 & 94.9 & 0.98 & 0.99 & 0.16 & 0.517 \\ \hline
\end{tabular}
}
\end{table*}

\subsection{Evaluation of effects of feature importance hiding}
\label{sec:er_hidden}

In order to demonstrate the effect of hiding feature importance, experimentation is presented on a classifier model with 50\% of the features intentionally eliminated from the classification process, in Table~\ref{tbl:hidden}. A random subspace model is chosen as the defender's model (50 Linear SVM, 50\% of features per model). A random subset of 50\% of the features are considered to be eliminated from the classification process. This is done by eliminating the features from the training dataset, and then training the defender's model on the reduced set of features. Any incoming probe is first truncated to the reduced feature set and then evaluated on the model. This makes the feature reduction strategy opaque to the external users. The results of the Anchor Points - High Attacks (AP - HC) attacks on this classifier is presented in Table~\ref{tbl:hidden}. The Effective Attack Rate (EAR) and the Adversarial Margin Density (AMD), are evaluated on this hidden feature classifier design and also a baseline classifier which uses all features in its model.

\begin{table}[t]
\centering
\caption{Effective Attack Rate (EAR) of models trained on  \textit{Available} and on  the \textit{Hidden} features, after AP-HC attack.}
\label{tbl:pd_results}
\begin{tabular}{|c|c|c|}
\hline
\multirow{2}{*}{Dataset} & \multicolumn{2}{c|}{EAR} \\ \cline{2-3} 
 & \begin{tabular}[c]{@{}c@{}}Available \\ Features\end{tabular} & \begin{tabular}[c]{@{}c@{}}Hidden \\ Features\end{tabular} \\ \hline
Synthetic & 0.99 & 0.67 \\ \hline
CAPTCHA & 0.99 & 0.65 \\ \hline
Phishing & 0.99 & 0.47 \\ \hline
Digits08 & 0.99 & 0.32 \\ \hline
Digits17 & 0.99 & 0.32 \\ \hline
\end{tabular}
\end{table}

It can be seen that the training accuracy is only marginally affected ($<$5\% difference at max), by the reduction of features from the models. This is a result of the presence of orthogonal information in the features of the training data. In case of a robust ensemble design, which incorporates all information in the model, the AMD is seen to drop as the attacks leverage the orthogonal information from the probes, to avoid regions of low certainty. However, for the hidden features approach, a significantly higher AMD value is seen in all cases (0.28, on average). This is because no amount of probing and cleansing will help the adversary in determining that the hidden features are important to the classification, and what exact values of the features they need to mimic. The EAR of all attacks is comparable to that of the robust classifier (Table~\ref{tbl:hidden}). However, the increased AMD ensures that the defender has an upper hand in the attack-defense cycle, as it can detect attacks and keep the training data clean for future retraining.  This makes hiding of feature importance an effective strategy in providing reactive security to adversarial drift. The notion of hiding feature importance suggests that not all information available at the training phase should be used at the same time, to ensure leverage over the adversary. A high AMD ensures that attacks can be detected and can be recovered from. The EAR of the two models: a) trained on the available set of features and b) trained on the hidden features, is shown in Table~\ref{tbl:pd_results}. It is seen that, while model trained on the set of available features is ultimately evaded by an adversary, the hidden features provide for a relatively unattacked set of information, which can be leveraged for retraining and continued usage, in a dynamic environment.

\section{Towards a \textit{Dynamic-Adversarial} mining approach}
\label{sec:da}

The analysis in this paper, takes initial steps towards understanding the security of machine learning as a holistic approach, combining proactive and reactive security measures. Long term effects of training time design decisions are analyzed. It was shown that restrictive classifier designs can be damaging for long term security, as they lead to data leakage and inseparability, after attacks. Based on the analysis, a counter-intuitive strategy was proposed, which advocates shielding of feature importance information. 

While existing works on \textit{Proactive} and on \textit{Reactive} security have largely been carried out in isolation from each other, there is a need for a combined dynamic view of the problem. Proactive steps can be taken in model development, to make subsequent detection and retraining easier. Similarly, honeypots and randomization, can be used effectively to delay the onset of attacks. The study of adversarial drift, as a specific case of the larger field of concept drift, is important to understand how adversarial behavior is affected by design decisions of the learned model. As adversarial drift is dependent on the defender's model (which the attacker is trying to reverse engineer and evade), it could be possible to train models to mislead adversary, and not just improve generalization. 

A holistic view of the security would understand that it is a cyclic process, where delaying attacks, detecting them and then recovering, are all equally important to meet the system's security goals. A dynamic security paradigm would take proactive measures to delay attacks, and to make their subsequent detection and fixing easier, as well. As such metrics like Attack delay, Difficulty to reverse engineer, Detect-ability of vulnerability of classifiers, and Recover-ability after attacks; are more useful to dynamic security, than the traditional metrics of accuracy, precision and recall. Also, dynamic security needs to operate as a \textit{never-ending} learning scheme, with efficient involvement of human in the loop, to provide expertise and retraining ability, from time to time. 

\begin{figure}[t]
  \centering
  \includegraphics[width=0.95\linewidth]{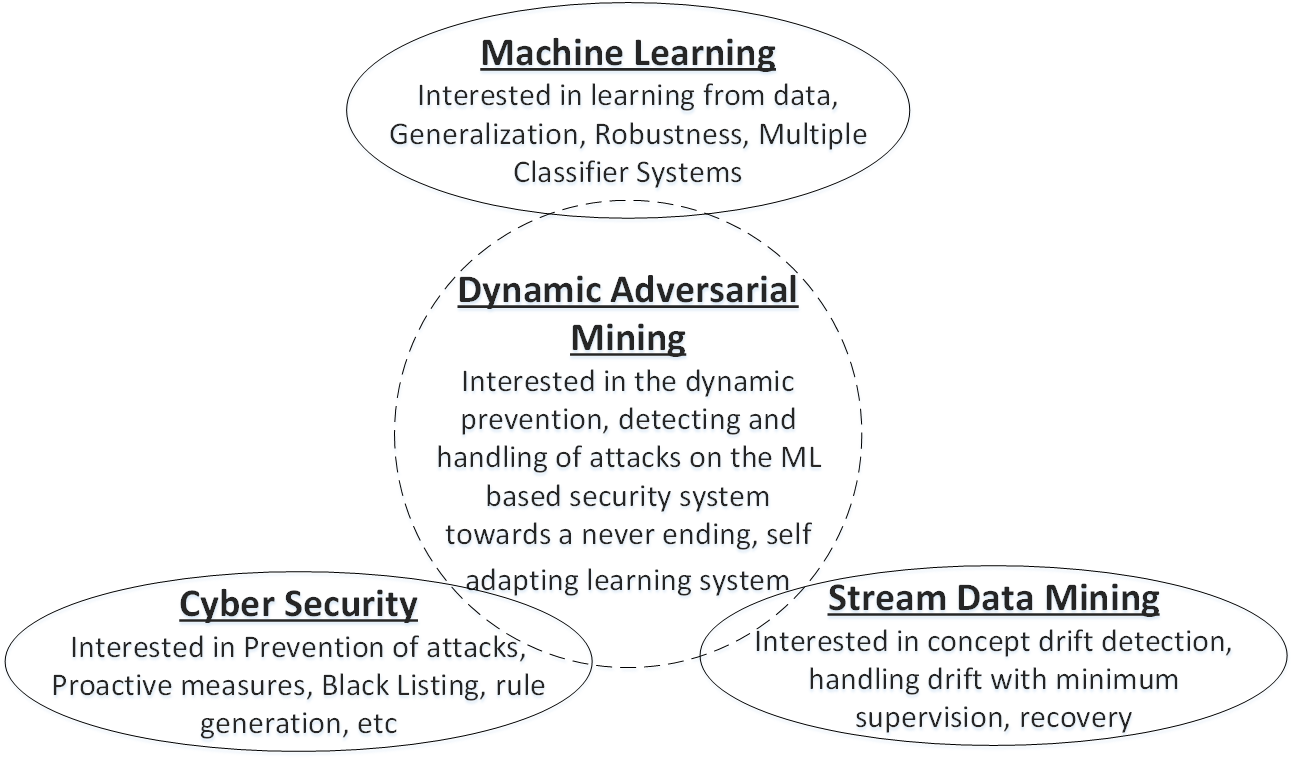}
   \caption{The field of \textit{Dynamic Adversarial Mining}, will derive from the concepts of Machine learning, Stream data mining and Cybersecurity.}
  \label{fig:intro2}
\end{figure}

The interdisciplinary field of \textit{Dynamic-Adversarial} mining (Figure~\ref{fig:intro2}), needs to incorporate lessons learned from: Machine Learning, Cybersecurity and Stream Data Mining; to rethink use of machine learning in security applications and develop systems which are \textit{`secure by design'} \citep{biggio2014pattern}. Some core ideas under the dynamic adversarial mining scheme are:

\begin{itemize}
\item Ability to leave feature space honeypots in the learned classifier at training time, to efficiently detect attacks using unlabeled data at test time.
\item Semi automated self aware algorithms, which can detect abnormal data distribution shifts, at test time.
\item Maintaining multiple backup models, which can provide prediction when the main model gets attacked.
\item Ability to use a stream labeling budget effectively and to distribute the budget appropriately, to detect and fix attacks. Thereby, managing human involvement in the process.
\item Understanding attack vulnerabilities on classifiers, from a purely data driven perspective, to effectively test the security measures employed.
\end{itemize}

This work takes steps towards an integrated approach of machine learning security, under the canopy of \textit{Dynamic-Adversarial} mining. Our work aims to encourage further awareness and interest in the need for a dynamic security paradigm, by bringing together ideas from the domains of  Streaming data and Cybersecurity.

\section{Conclusion and future work}
\label{sec:conclusion}

Although the naive application of machine learning, has found early success in many domains, its abilities and vulnerabilities are still not well understood. To secure machine learning and to make it usable in a long term dynamic environment, it is necessary to adopt a holistic view in its analysis. In this paper, the shortcomings in popular classifier design strategies is highlighted, as they tend to approach security from a static perspective. Approaches such as restrictive one class classifiers, robust feature bagged ensembles and randomization based ensembles, are all found to leak excessive information to the adversary. This excessive information can be leveraged by an intelligent adversary, who can launch attacks, such that future detection and recovery from attacks is made difficult. Such vulnerabilities are introduced into the machine learning model, due to the widely accepted philosophy of incorporating excessive information into the defender's model, in the hopes of making it more secure. However, a dynamic analysis of this philosophy shows that it is not applicable as a long term solution.

Based on the analysis of the static solutions, a novel \textit{Dynamic-Adversarial} aware solution is proposed. This methodology intelligently shields a subset of the features, from being probed and leaked to the adversary, by hiding their impact on the classification task. This resulted in adversary's with a misguided sense of confidence and also enables better attack detection and recover. The proposed methodology serves as a blueprint for further research in the area of \textit{Dynamic-Adversarial} mining, where a holistic approach to security will lead to long term benefits for the defender. Future work will concentrate on developing secure methods for retraining classifiers, with the new labeled data. Also, the development of an adversarial aware concept drift handling methodology, is an area warranting further research.





\section*{References}

\bibliographystyle{elsarticle-harv} 
\bibliography{references}

\begin{thebibliography}{59}
\expandafter\ifx\csname natexlab\endcsname\relax\def\natexlab#1{#1}\fi
\expandafter\ifx\csname url\endcsname\relax
  \def\url#1{\texttt{#1}}\fi
\expandafter\ifx\csname urlprefix\endcsname\relax\def\urlprefix{URL }\fi

\bibitem[{Ahn et~al.(2007)Ahn, Moon, Fazzari, Lim, Chen, and
  Kodell}]{ahn2007classification}
Ahn, H., Moon, H., Fazzari, M.~J., Lim, N., Chen, J.~J., Kodell, R.~L., 2007.
  Classification by ensembles from random partitions of high-dimensional data.
  Computational Statistics \& Data Analysis 51~(12), 6166--6179.

\bibitem[{Bach and Maloof(2008)}]{bach2008paired}
Bach, S.~H., Maloof, M.~A., 2008. Paired learners for concept drift. In: Eighth
  IEEE International Conference on Data Mining (ICDM). IEEE, pp. 23--32.

\bibitem[{Baena-Garc{\i}a et~al.(2006)Baena-Garc{\i}a, del Campo-{\'A}vila,
  Fidalgo, Bifet, Gavalda, and Morales-Bueno}]{baena2006early}
Baena-Garc{\i}a, M., del Campo-{\'A}vila, J., Fidalgo, R., Bifet, A., Gavalda,
  R., Morales-Bueno, R., 2006. Early drift detection method. In: Fourth
  international workshop on knowledge discovery from data streams. Vol.~6. pp.
  77--86.

\bibitem[{Barreno et~al.(2010)Barreno, Nelson, Joseph, and
  Tygar}]{barreno2010security}
Barreno, M., Nelson, B., Joseph, A.~D., Tygar, J., 2010. The security of
  machine learning. Machine Learning 81~(2), 121--148.

\bibitem[{Bifet and Gavalda(2007)}]{bifet2007learning}
Bifet, A., Gavalda, R., 2007. Learning from time-changing data with adaptive
  windowing. In: SDM. Vol.~7. SIAM.

\bibitem[{Bryll et~al.(2003)Bryll, Gutierrez-Osuna, and Quek}]{bry}
Bryll, R., Gutierrez-Osuna, R., Quek, F., 2003. Attribute bagging: improving
  accuracy of classifier ensembles by using random feature subsets. Pattern
  recognition 36~(6), 1291--1302.

\bibitem[{Burnap and Williams(2016)}]{burnap2016us}
Burnap, P., Williams, M.~L., 2016. Us and them: identifying cyber hate on
  twitter across multiple protected characteristics. EPJ Data Science 5~(1), 1.

\bibitem[{Cha(2007)}]{cha2007comprehensive}
Cha, S.-H., 2007. Comprehensive survey on distance/similarity measures between
  probability density functions. City 1~(2), 1.

\bibitem[{Chang and Lin(2011)}]{chang2011libsvm}
Chang, C.-C., Lin, C.-J., 2011. Libsvm: a library for support vector machines.
  ACM Transactions on Intelligent Systems and Technology (TIST) 2~(3), 27.

\bibitem[{Chinavle et~al.(2009)Chinavle, Kolari, Oates, and
  Finin}]{chinavle2009ensembles}
Chinavle, D., Kolari, P., Oates, T., Finin, T., 2009. Ensembles in adversarial
  classification for spam. In: Proceedings of the 18th ACM conference on
  Information and knowledge management. ACM, pp. 2015--2018.

\bibitem[{Cover and Hart(1967)}]{cover1967nearest}
Cover, T.~M., Hart, P.~E., 1967. Nearest neighbor pattern classification. IEEE
  Transactions on Information Theory 13~(1), 21--27.

\bibitem[{da~Costa et~al.(2016)da~Costa, Rios, and de~Mello}]{da2016using}
da~Costa, F., Rios, R., de~Mello, R., 2016. Using dynamical systems tools to
  detect concept drift in data streams. Expert Systems with Applications 60,
  39--50.

\bibitem[{Dietterich(2000)}]{dietterich2000ensemble}
Dietterich, T.~G., 2000. Ensemble methods in machine learning. In: Multiple
  classifier systems. Springer, pp. 1--15.

\bibitem[{Ditzler and Polikar(2011)}]{ditzler2011hellinger}
Ditzler, G., Polikar, R., 2011. Hellinger distance based drift detection for
  nonstationary environments. In: IEEE Symposium on Computational Intelligence
  in Dynamic and Uncertain Environments (CIDUE). IEEE, pp. 41--48.

\bibitem[{Dredze et~al.(2010)Dredze, Oates, and Piatko}]{dredze2010we}
Dredze, M., Oates, T., Piatko, C., 2010. We're not in kansas anymore: detecting
  domain changes in streams. In: Proceedings of the 2010 Conference on
  Empirical Methods in Natural Language Processing. Association for
  Computational Linguistics, pp. 585--595.

\bibitem[{Dries and R{\"u}ckert(2009)}]{dries2009adaptive}
Dries, A., R{\"u}ckert, U., 2009. Adaptive concept drift detection. Statistical
  Analysis and Data Mining 2~(5-6), 311--327.

\bibitem[{Duch et~al.(2004)Duch, Wieczorek, Biesiada, and
  Blachnik}]{duch2004comparison}
Duch, W., Wieczorek, T., Biesiada, J., Blachnik, M., 2004. Comparison of
  feature ranking methods based on information entropy. In: IEEE International
  Joint Conference on Neural Networks. Vol.~2. IEEE, pp. 1415--1419.

\bibitem[{Faria et~al.(2013)Faria, Gama, and Carvalho}]{faria2013novelty}
Faria, E.~R., Gama, J., Carvalho, A.~C., 2013. Novelty detection algorithm for
  data streams multi-class problems. In: Proceedings of the 28th Annual ACM
  Symposium on Applied Computing. ACM, pp. 795--800.

\bibitem[{Farid et~al.(2013)Farid, Zhang, Hossain, Rahman, Strachan, Sexton,
  and Dahal}]{farid2013adaptive}
Farid, D.~M., Zhang, L., Hossain, A., Rahman, C.~M., Strachan, R., Sexton, G.,
  Dahal, K., 2013. An adaptive ensemble classifier for mining concept drifting
  data streams. Expert Systems with Applications 40~(15), 5895--5906.

\bibitem[{Gama et~al.(2004)Gama, Medas, Castillo, and
  Rodrigues}]{gama2004learning}
Gama, J., Medas, P., Castillo, G., Rodrigues, P., 2004. Learning with drift
  detection. In: Advances in artificial intelligence--SBIA 2004. Springer, pp.
  286--295.

\bibitem[{Gao et~al.(2007)Gao, Fan, and Han}]{gao2007appropriate}
Gao, J., Fan, W., Han, J., 2007. On appropriate assumptions to mine data
  streams: Analysis and practice. In: Seventh IEEE International Conference on
  Data Mining (ICDM). IEEE, pp. 143--152.

\bibitem[{Goncalves et~al.(2014)Goncalves, de~Carvalho~Santos, Barros, and
  Vieira}]{goncalves2014comparative}
Goncalves, P.~M., de~Carvalho~Santos, S.~G., Barros, R.~S., Vieira, D.~C.,
  2014. A comparative study on concept drift detectors. Expert Systems with
  Applications 41~(18), 8144--8156.

\bibitem[{Harel et~al.(2014)Harel, Mannor, El-Yaniv, and
  Crammer}]{harel2014concept}
Harel, M., Mannor, S., El-Yaniv, R., Crammer, K., 2014. Concept drift detection
  through resampling. In: Proceedings of the 31st International Conference on
  Machine Learning (ICML-14). pp. 1009--1017.

\bibitem[{Haussler(1990)}]{haussler1990probably}
Haussler, D., 1990. Probably approximately correct learning. University of
  California, Santa Cruz, Computer Research Laboratory.

\bibitem[{Hayat and Hashemi(2010)}]{hayat2010dct}
Hayat, M.~Z., Hashemi, M.~R., 2010. A dct based approach for detecting novelty
  and concept drift in data streams. In: International Conference of Soft
  Computing and Pattern Recognition (SoCPaR). IEEE, pp. 373--378.

\bibitem[{Katakis et~al.(2009)Katakis, Tsoumakas, Banos, Bassiliades, and
  Vlahavas}]{katakis2009adaptive}
Katakis, I., Tsoumakas, G., Banos, E., Bassiliades, N., Vlahavas, I., 2009. An
  adaptive personalized news dissemination system. Journal of Intelligent
  Information Systems 32~(2), 191--212.

\bibitem[{Katakis et~al.(2010)Katakis, Tsoumakas, and
  Vlahavas}]{katakis2010tracking}
Katakis, I., Tsoumakas, G., Vlahavas, I., 2010. Tracking recurring contexts
  using ensemble classifiers: an application to email filtering. Knowledge and
  Information Systems 22~(3), 371--391.

\bibitem[{Kohavi et~al.(1995)}]{kohavi1995study}
Kohavi, R., et~al., 1995. A study of cross-validation and bootstrap for
  accuracy estimation and model selection. In: Ijcai. Vol.~14. pp. 1137--1145.

\bibitem[{Krempl et~al.(2014)Krempl, {\v{Z}}liobaite, Brzezi{\'n}ski,
  H{\"u}llermeier, Last, Lemaire, Noack, Shaker, Sievi, Spiliopoulou,
  et~al.}]{krempl2014open}
Krempl, G., {\v{Z}}liobaite, I., Brzezi{\'n}ski, D., H{\"u}llermeier, E., Last,
  M., Lemaire, V., Noack, T., Shaker, A., Sievi, S., Spiliopoulou, M., et~al.,
  2014. Open challenges for data stream mining research. ACM SIGKDD
  explorations newsletter 16~(1), 1--10.

\bibitem[{Kuncheva(2008)}]{kuncheva2008classifier}
Kuncheva, L.~I., 2008. Classifier ensembles for detecting concept change in
  streaming data: Overview and perspectives. In: 2nd Workshop SUEMA. Vol. 2008.
  pp. 5--10.

\bibitem[{Kuncheva and Faithfull(2014)}]{kuncheva2014pca}
Kuncheva, L.~I., Faithfull, W.~J., 2014. Pca feature extraction for change
  detection in multidimensional unlabeled data. IEEE Transactions on Neural
  Networks and Learning Systems 25~(1), 69--80.

\bibitem[{Lee and Magoules(2012)}]{lee2012detection}
Lee, J., Magoules, F., 2012. Detection of concept drift for learning from
  stream data. In: IEEE 14th International Conference on High Performance
  Computing and Communication \& 2012 IEEE 9th International Conference on
  Embedded Software and Systems (HPCC-ICESS). IEEE, pp. 241--245.

\bibitem[{Lichman(2013)}]{Lichman:2013}
Lichman, M., 2013. {UCI} machine learning repository.
\newline\urlprefix\url{http://archive.ics.uci.edu/ml}

\bibitem[{Lindstrom et~al.(2010)Lindstrom, Delany, and
  Mac~Namee}]{lindstrom2010handling}
Lindstrom, P., Delany, S.~J., Mac~Namee, B., 2010. Handling concept drift in a
  text data stream constrained by high labelling cost. In: Twenty-Third
  International FLAIRS Conference.

\bibitem[{Lindstrom et~al.(2013)Lindstrom, Mac~Namee, and
  Delany}]{lindstrom2013drift}
Lindstrom, P., Mac~Namee, B., Delany, S.~J., 2013. Drift detection using
  uncertainty distribution divergence. Evolving Systems 4~(1), 13--25.

\bibitem[{Lughofer et~al.(2016)Lughofer, Weigl, Heidl, Eitzinger, and
  Radauer}]{lughofer2016recognizing}
Lughofer, E., Weigl, E., Heidl, W., Eitzinger, C., Radauer, T., 2016.
  Recognizing input space and target concept drifts in data streams with
  scarcely labeled and unlabelled instances. Information Sciences 355,
  127--151.

\bibitem[{Masud et~al.(2011)Masud, Gao, Khan, Han, and
  Thuraisingham}]{masud2011classification}
Masud, M.~M., Gao, J., Khan, L., Han, J., Thuraisingham, B., 2011.
  Classification and novel class detection in concept-drifting data streams
  under time constraints. IEEE TKDE 23~(6), 859--874.

\bibitem[{Nishida and Yamauchi(2007)}]{nishida2007detecting}
Nishida, K., Yamauchi, K., 2007. Detecting concept drift using statistical
  testing. In: Discovery Science. Springer, pp. 264--269.

\bibitem[{Page(1954)}]{page1954continuous}
Page, E., 1954. Continuous inspection schemes. Biometrika 41~(1/2), 100--115.

\bibitem[{Pedregosa et~al.(2011)Pedregosa, Varoquaux, Gramfort, Michel,
  Thirion, Grisel, Blondel, Prettenhofer, Weiss, Dubourg, Vanderplas, Passos,
  Cournapeau, Brucher, Perrot, and Duchesnay}]{scikit-learn}
Pedregosa, F., Varoquaux, G., Gramfort, A., Michel, V., Thirion, B., Grisel,
  O., Blondel, M., Prettenhofer, P., Weiss, R., Dubourg, V., Vanderplas, J.,
  Passos, A., Cournapeau, D., Brucher, M., Perrot, M., Duchesnay, E., 2011.
  Scikit-learn: Machine learning in {P}ython. Journal of Machine Learning
  Research 12, 2825--2830.

\bibitem[{Qahtan et~al.(2015)Qahtan, Alharbi, Wang, and Zhang}]{qahtan2015pca}
Qahtan, A.~A., Alharbi, B., Wang, S., Zhang, X., 2015. A pca-based change
  detection framework for multidimensional data streams: Change detection in
  multidimensional data streams. In: Proc. of the 21th ACM SIGKDD ICKDDM. ACM,
  pp. 935--944.

\bibitem[{Quinlan(1993)}]{quinlan1993c4}
Quinlan, J.~R., 1993. C4. 5: Programming for machine learning. Morgan
  Kauffmann.

\bibitem[{Ross et~al.(2012)Ross, Adams, Tasoulis, and
  Hand}]{ross2012exponentially}
Ross, G.~J., Adams, N.~M., Tasoulis, D.~K., Hand, D.~J., 2012. Exponentially
  weighted moving average charts for detecting concept drift. Pattern
  Recognition Letters 33~(2), 191--198.

\bibitem[{Ryu et~al.(2012)Ryu, Kantardzic, Kim, and Khil}]{ryu2012efficient}
Ryu, J.~W., Kantardzic, M.~M., Kim, M.-W., Khil, A.~R., 2012. An efficient
  method of building an ensemble of classifiers in streaming data. In: Big data
  analytics. Springer, pp. 122--133.

\bibitem[{Schmidt et~al.(1995)Schmidt, Siegel, and
  Srinivasan}]{schmidt1995chernoff}
Schmidt, J.~P., Siegel, A., Srinivasan, A., 1995. Chernoff-hoeffding bounds for
  applications with limited independence. SIAM Journal on Discrete Mathematics
  8~(2), 223--250.

\bibitem[{Sethi and Kantardzic(2015)}]{sethi2015don}
Sethi, T.~S., Kantardzic, M., 2015. Don't pay for validation: Detecting drifts
  from unlabeled data using margin density. Procedia Computer Science 53,
  103--112.

\bibitem[{Sethi et~al.(2016{\natexlab{a}})Sethi, Kantardzic, and
  Arabmakki}]{tsethi}
Sethi, T.~S., Kantardzic, M., Arabmakki, E., 2016{\natexlab{a}}. Monitoring
  classification blindspots to detect drifts from unlabeled data. In: 17th IEEE
  International Conference on Information Reuse and Integration (IRI). IEEE.

\bibitem[{Sethi et~al.(2016{\natexlab{b}})Sethi, Kantardzic, and
  Hu}]{sethi2016grid}
Sethi, T.~S., Kantardzic, M., Hu, H., 2016{\natexlab{b}}. A grid density based
  framework for classifying streaming data in the presence of concept drift.
  Journal of Intelligent Information Systems 46~(1), 179--211.

\bibitem[{Settles(2012)}]{settles2010active}
Settles, B., 2012. Active learning. Synthesis Lectures on Artificial
  Intelligence and Machine Learning 6~(1), 1--114.

\bibitem[{Skurichina and Duin(2002)}]{skurichina2002bagging}
Skurichina, M., Duin, R.~P., 2002. Bagging, boosting and the random subspace
  method for linear classifiers. Pattern Analysis \& Applications 5~(2),
  121--135.

\bibitem[{Smutz and Stavrou(2016)}]{smutz2016tree}
Smutz, C., Stavrou, A., 2016. When a tree falls: Using diversity in ensemble
  classifiers to identify evasion in malware detectors. In: NDSS Symposium.

\bibitem[{Sobhani and Beigy(2011)}]{sobhani2011new}
Sobhani, P., Beigy, H., 2011. New drift detection method for data streams.
  Springer.

\bibitem[{Spinosa et~al.(2007)Spinosa, de~Leon F~de Carvalho, and
  Gama}]{spinosa2007olindda}
Spinosa, E.~J., de~Leon F~de Carvalho, A.~P., Gama, J., 2007. Olindda: A
  cluster-based approach for detecting novelty and concept drift in data
  streams. In: Proceedings of the 2007 ACM symposium on Applied computing. ACM,
  pp. 448--452.

\bibitem[{Tavallaee et~al.(2009)Tavallaee, Bagheri, Lu, and
  Ghorbani}]{tavallaee2009detailed}
Tavallaee, M., Bagheri, E., Lu, W., Ghorbani, A.-A., 2009. A detailed analysis
  of the kdd cup 99 data set. In: Proceedings of the Second IEEE Symposium on
  Computational Intelligence for Security and Defence Applications 2009.

\bibitem[{Wang(2015)}]{wang2015robust}
Wang, F., 2015. Robust and adversarial data mining. Sydney Digital Theses (Open
  Access).

\bibitem[{Wang and Abraham(2015)}]{wang2015concept}
Wang, H., Abraham, Z., 2015. Concept drift detection for streaming data. In:
  International Joint Conference on Neural Networks (IJCNN). IEEE, pp. 1--9.

\bibitem[{Wu et~al.(2014)Wu, Zhu, Wu, and Ding}]{wu2014data}
Wu, X., Zhu, X., Wu, G.-Q., Ding, W., 2014. Data mining with big data.
  Knowledge and Data Engineering, IEEE Transactions on 26~(1), 97--107.

\bibitem[{Zliobaite(2010)}]{zliobaite2010change}
Zliobaite, I., 2010. Change with delayed labeling: when is it detectable? In:
  IEEE International Conference on Data Mining Workshops (ICDMW). IEEE, pp.
  843--850.

\bibitem[{Zliobaite et~al.(2014)Zliobaite, Bifet, Pfahringer, and
  Holmes}]{zliobaite2014active}
Zliobaite, I., Bifet, A., Pfahringer, B., Holmes, G., 2014. Active learning
  with drifting streaming data. IEEE Transactions on Neural Networks and
  Learning Systems 25~(1), 27--39.

\end{thebibliography}


\begin{thebibliography}{68}
\expandafter\ifx\csname natexlab\endcsname\relax\def\natexlab#1{#1}\fi
\expandafter\ifx\csname url\endcsname\relax
  \def\url#1{\texttt{#1}}\fi
\expandafter\ifx\csname urlprefix\endcsname\relax\def\urlprefix{URL }\fi

\bibitem[{Abramson(2015)}]{abramson2015toward}
Abramson, M., 2015. Toward adversarial online learning and the science of
  deceptive machines. In: 2015 AAAI Fall Symposium Series.

\bibitem[{Alabdulmohsin et~al.(2014)Alabdulmohsin, Gao, and
  Zhang}]{alabdulmohsin2014adding}
Alabdulmohsin, I.~M., Gao, X., Zhang, X., 2014. Adding robustness to support
  vector machines against adversarial reverse engineering. In: Proceedings of
  the 23rd ACM International Conference on Conference on Information and
  Knowledge Management. ACM, pp. 231--240.

\bibitem[{Barreno et~al.(2010)Barreno, Nelson, Joseph, and
  Tygar}]{barreno2010security}
Barreno, M., Nelson, B., Joseph, A.~D., Tygar, J., 2010. The security of
  machine learning. Machine Learning 81~(2), 121--148.

\bibitem[{Barreno et~al.(2006)Barreno, Nelson, Sears, Joseph, and
  Tygar}]{barreno2006can}
Barreno, M., Nelson, B., Sears, R., Joseph, A.~D., Tygar, J.~D., 2006. Can
  machine learning be secure? In: Proceedings of the 2006 ACM Symposium on
  Information, computer and communications security. ACM, pp. 16--25.

\bibitem[{Barth et~al.(2012)Barth, Rubinstein, Sundararajan, Mitchell, Song,
  and Bartlett}]{barth2012learning}
Barth, A., Rubinstein, B.~I., Sundararajan, M., Mitchell, J.~C., Song, D.,
  Bartlett, P.~L., 2012. A learning-based approach to reactive security. IEEE
  Transactions on Dependable and Secure Computing 9~(4), 482--493.

\bibitem[{Biggio et~al.(2015)Biggio, Corona, He, Chan, Giacinto, Yeung, and
  Roli}]{biggio2015one}
Biggio, B., Corona, I., He, Z.-M., Chan, P.~P., Giacinto, G., Yeung, D.~S.,
  Roli, F., 2015. One-and-a-half-class multiple classifier systems for secure
  learning against evasion attacks at test time. In: Multiple Classifier
  Systems. Springer, pp. 168--180.

\bibitem[{Biggio et~al.(2013)Biggio, Corona, Maiorca, Nelson, {\v{S}}rndi{\'c},
  Laskov, Giacinto, and Roli}]{biggio2013evasion}
Biggio, B., Corona, I., Maiorca, D., Nelson, B., {\v{S}}rndi{\'c}, N., Laskov,
  P., Giacinto, G., Roli, F., 2013. Evasion attacks against machine learning at
  test time. In: Machine Learning and Knowledge Discovery in Databases.
  Springer, pp. 387--402.

\bibitem[{Biggio et~al.(2008)Biggio, Fumera, and Roli}]{biggio2008adversarial}
Biggio, B., Fumera, G., Roli, F., 2008. Adversarial pattern classification
  using multiple classifiers and randomisation. In: Structural, Syntactic, and
  Statistical Pattern Recognition. Springer, pp. 500--509.

\bibitem[{Biggio et~al.(2010{\natexlab{a}})Biggio, Fumera, and
  Roli}]{biggio2010multiple}
Biggio, B., Fumera, G., Roli, F., 2010{\natexlab{a}}. Multiple classifier
  systems for robust classifier design in adversarial environments.
  International Journal of Machine Learning and Cybernetics 1~(1-4), 27--41.

\bibitem[{Biggio et~al.(2010{\natexlab{b}})Biggio, Fumera, and
  Roli}]{biggio2010multipleattack}
Biggio, B., Fumera, G., Roli, F., 2010{\natexlab{b}}. Multiple classifier
  systems under attack. In: Multiple Classifier Systems. Springer, pp. 74--83.

\bibitem[{Biggio et~al.(2014{\natexlab{a}})Biggio, Fumera, and
  Roli}]{biggio2014pattern}
Biggio, B., Fumera, G., Roli, F., 2014{\natexlab{a}}. Pattern recognition
  systems under attack: Design issues and research challenges. International
  Journal of Pattern Recognition and Artificial Intelligence 28~(07), 1460002.

\bibitem[{Biggio et~al.(2014{\natexlab{b}})Biggio, Fumera, and
  Roli}]{biggio2014security}
Biggio, B., Fumera, G., Roli, F., 2014{\natexlab{b}}. Security evaluation of
  pattern classifiers under attack. IEEE Transactions on Knowledge and Data
  Engineering 26~(4), 984--996.

\bibitem[{Bryll et~al.(2003)Bryll, Gutierrez-Osuna, and Quek}]{bry}
Bryll, R., Gutierrez-Osuna, R., Quek, F., 2003. Attribute bagging: improving
  accuracy of classifier ensembles by using random feature subsets. Pattern
  recognition 36~(6), 1291--1302.

\bibitem[{Brzezinski and Stefanowski(2014)}]{brzezinski2014reacting}
Brzezinski, D., Stefanowski, J., 2014. Reacting to different types of concept
  drift: The accuracy updated ensemble algorithm. IEEE Transactions on Neural
  Networks and Learning Systems 25~(1), 81--94.

\bibitem[{Carlini et~al.(2016)Carlini, Mishra, Vaidya, Zhang, Sherr, Shields,
  Wagner, and Zhou}]{carlini2016hidden}
Carlini, N., Mishra, P., Vaidya, T., Zhang, Y., Sherr, M., Shields, C., Wagner,
  D., Zhou, W., 2016. Hidden voice commands. In: USENIX Security Symposium. pp.
  513--530.

\bibitem[{Chang and Lin(2011)}]{chang2011libsvm}
Chang, C.-C., Lin, C.-J., 2011. Libsvm: a library for support vector machines.
  ACM Transactions on Intelligent Systems and Technology (TIST) 2~(3), 27.

\bibitem[{Chinavle et~al.(2009)Chinavle, Kolari, Oates, and
  Finin}]{chinavle2009ensembles}
Chinavle, D., Kolari, P., Oates, T., Finin, T., 2009. Ensembles in adversarial
  classification for spam. In: Proceedings of the 18th ACM conference on
  Information and knowledge management. ACM, pp. 2015--2018.

\bibitem[{Colbaugh and Glass(2012{\natexlab{a}})}]{colbaugh2012predictability}
Colbaugh, R., Glass, K., 2012{\natexlab{a}}. Predictability-oriented defense
  against adaptive adversaries. In: IEEE International Conference on Systems,
  Man, and Cybernetics (SMC). IEEE, pp. 2721--2727.

\bibitem[{Colbaugh and Glass(2012{\natexlab{b}})}]{colbaugh2012predictive}
Colbaugh, R., Glass, K., 2012{\natexlab{b}}. Predictive defense against
  evolving adversaries. In: IEEE International Conference on Intelligence and
  Security Informatics (ISI). IEEE, pp. 18--23.

\bibitem[{Dalvi et~al.(2004)Dalvi, Domingos, Sanghai, Verma,
  et~al.}]{dalvi2004adversarial}
Dalvi, N., Domingos, P., Sanghai, S., Verma, D., et~al., 2004. Adversarial
  classification. In: Proceedings of the tenth ACM SIGKDD international
  conference on Knowledge discovery and data mining. ACM, pp. 99--108.

\bibitem[{Ditzler et~al.(2015)Ditzler, Roveri, Alippi, and
  Polikar}]{ditzler2015learning}
Ditzler, G., Roveri, M., Alippi, C., Polikar, R., 2015. Learning in
  nonstationary environments: A survey. Computational Intelligence Magazine
  10~(4), 12--25.

\bibitem[{D’Souza(2014)}]{d2014avatar}
D’Souza, D.~F., 2014. Avatar captcha: telling computers and humans apart via
  face classification and mouse dynamics.

\bibitem[{Gama et~al.(2014)Gama, {\v{Z}}liobait{\.e}, Bifet, Pechenizkiy, and
  Bouchachia}]{gama2014survey}
Gama, J., {\v{Z}}liobait{\.e}, I., Bifet, A., Pechenizkiy, M., Bouchachia, A.,
  2014. A survey on concept drift adaptation. ACM Computing Surveys (CSUR)
  46~(4), 44.

\bibitem[{Globerson and Roweis(2006)}]{globerson2006nightmare}
Globerson, A., Roweis, S., 2006. Nightmare at test time: robust learning by
  feature deletion. In: Proceedings of the 23rd international conference on
  Machine learning. ACM, pp. 353--360.

\bibitem[{Hardt et~al.(2016)Hardt, Megiddo, Papadimitriou, and
  Wootters}]{hardt2016strategic}
Hardt, M., Megiddo, N., Papadimitriou, C., Wootters, M., 2016. Strategic
  classification. In: Proceedings of the 2016 ACM Conference on Innovations in
  Theoretical Computer Science. ACM, pp. 111--122.

\bibitem[{Henke et~al.(2015)Henke, Souto, and dos Santos}]{henke2015analysis}
Henke, M., Souto, E., dos Santos, E.~M., 2015. Analysis of the evolution of
  features in classification problems with concept drift: Application to spam
  detection. In: IFIP/IEEE International Symposium on Integrated Network
  Management (IM). IEEE, pp. 874--877.

\bibitem[{Ho(1998)}]{ho1998random}
Ho, T.~K., 1998. The random subspace method for constructing decision forests.
  IEEE Transactions on Pattern Analysis and Machine Intelligence 20~(8),
  832--844.

\bibitem[{Hosseini et~al.(2017)Hosseini, Kannan, Zhang, and
  Poovendran}]{hosseini2017deceiving}
Hosseini, H., Kannan, S., Zhang, B., Poovendran, R., 2017. Deceiving google's
  perspective api built for detecting toxic comments. arXiv preprint
  arXiv:1702.08138.

\bibitem[{Huang et~al.(2011)Huang, Joseph, Nelson, Rubinstein, and
  Tygar}]{huang2011adversarial}
Huang, L., Joseph, A.~D., Nelson, B., Rubinstein, B.~I., Tygar, J., 2011.
  Adversarial machine learning. In: Proceedings of the 4th ACM workshop on
  Security and artificial intelligence. ACM, pp. 43--58.

\bibitem[{Kantchelian et~al.(2013)Kantchelian, Afroz, Huang, Islam, Miller,
  Tschantz, Greenstadt, Joseph, and Tygar}]{kantchelian2013approaches}
Kantchelian, A., Afroz, S., Huang, L., Islam, A.~C., Miller, B., Tschantz,
  M.~C., Greenstadt, R., Joseph, A.~D., Tygar, J., 2013. Approaches to
  adversarial drift. In: Proceedings of the 2013 ACM workshop on Artificial
  intelligence and security. ACM, pp. 99--110.

\bibitem[{Kerckhoffs(1883)}]{kerckhoffs1883cryptographie}
Kerckhoffs, A., 1883. La cryptographie militaire (military cryptography),‖ j.
  Sciences Militaires (J. Military Science, in French).

\bibitem[{Ko{\l}cz and Teo(2009)}]{kolcz2009feature}
Ko{\l}cz, A., Teo, C.~H., 2009. Feature weighting for improved classifier
  robustness. In: CEAS’09: sixth conference on email and anti-spam.

\bibitem[{Kuncheva(2008)}]{kuncheva2008classifier}
Kuncheva, L.~I., 2008. Classifier ensembles for detecting concept change in
  streaming data: Overview and perspectives. In: 2nd Workshop SUEMA. Vol. 2008.
  pp. 5--10.

\bibitem[{Lee et~al.(2010)Lee, Caverlee, and Webb}]{lee2010uncovering}
Lee, K., Caverlee, J., Webb, S., 2010. Uncovering social spammers: social
  honeypots+ machine learning. In: Proceedings of the 33rd international ACM
  SIGIR conference on Research and development in information retrieval. ACM,
  pp. 435--442.

\bibitem[{Lichman(2013)}]{Lichman:2013}
Lichman, M., 2013. {UCI} machine learning repository.
\newline\urlprefix\url{http://archive.ics.uci.edu/ml}

\bibitem[{Liu et~al.(2012)Liu, Chawla, Bailey, Leckie, and
  Ramamohanarao}]{liu2012efficient}
Liu, W., Chawla, S., Bailey, J., Leckie, C., Ramamohanarao, K., 2012. An
  efficient adversarial learning strategy for constructing robust
  classification boundaries. In: AI 2012: Advances in Artificial Intelligence.
  Springer, pp. 649--660.

\bibitem[{Lowd and Meek(2005)}]{lowd2005adversarial}
Lowd, D., Meek, C., 2005. Adversarial learning. In: Proceedings of the eleventh
  ACM SIGKDD international conference on Knowledge discovery in data mining.
  ACM, pp. 641--647.

\bibitem[{Miller et~al.(2014)Miller, Kantchelian, Afroz, Bachwani, Dauber,
  Huang, Tschantz, Joseph, and Tygar}]{miller2014adversarial}
Miller, B., Kantchelian, A., Afroz, S., Bachwani, R., Dauber, E., Huang, L.,
  Tschantz, M.~C., Joseph, A.~D., Tygar, J.~D., 2014. Adversarial active
  learning. In: Proceedings of the 2014 Workshop on Artificial Intelligent and
  Security Workshop. ACM, pp. 3--14.

\bibitem[{Minku and Yao(2012)}]{minku2012ddd}
Minku, L.~L., Yao, X., 2012. Ddd: A new ensemble approach for dealing with
  concept drift. IEEE Transactions on Knowledge and Data Engineering 24~(4),
  619--633.

\bibitem[{Mrdovic and Perunicic(2008)}]{mrdovic2008kerckhoffs}
Mrdovic, S., Perunicic, B., 2008. Kerckhoffs' principle for intrusion
  detection. In: The 13th International Telecommunications Network Strategy and
  Planning Symposium, 2008. IEEE, pp. 1--8.

\bibitem[{Mthembu and Marwala(2008)}]{mthembu2008note}
Mthembu, L., Marwala, T., 2008. A note on the separability index. arXiv
  preprint arXiv:0812.1107.

\bibitem[{Onoda and Kiuchi(2012)}]{onoda2012analysis}
Onoda, T., Kiuchi, M., 2012. Analysis of intrusion detection in control system
  communication based on outlier detection with one-class classifiers. In:
  Neural Information Processing. Springer, pp. 275--282.

\bibitem[{Papernot et~al.(2016{\natexlab{a}})Papernot, McDaniel, and
  Goodfellow}]{papernot2016transferability}
Papernot, N., McDaniel, P., Goodfellow, I., 2016{\natexlab{a}}. Transferability
  in machine learning: from phenomena to black-box attacks using adversarial
  samples. arXiv preprint arXiv:1605.07277.

\bibitem[{Papernot et~al.(2017)Papernot, McDaniel, Goodfellow, Jha, Celik, and
  Swami}]{papernot2017practical}
Papernot, N., McDaniel, P., Goodfellow, I., Jha, S., Celik, Z.~B., Swami, A.,
  2017. Practical black-box attacks against machine learning. In: Proceedings
  of the 2017 ACM on Asia Conference on Computer and Communications Security.
  ACM, pp. 506--519.

\bibitem[{Papernot et~al.(2016{\natexlab{b}})Papernot, McDaniel, Jha,
  Fredrikson, Celik, and Swami}]{papernot2016limitations}
Papernot, N., McDaniel, P., Jha, S., Fredrikson, M., Celik, Z.~B., Swami, A.,
  2016{\natexlab{b}}. The limitations of deep learning in adversarial settings.
  In: 2016 IEEE European Symposium on Security and Privacy (EuroS\&P). IEEE,
  pp. 372--387.

\bibitem[{Papernot et~al.(2016{\natexlab{c}})Papernot, McDaniel, Sinha, and
  Wellman}]{papernot2016towards}
Papernot, N., McDaniel, P., Sinha, A., Wellman, M., 2016{\natexlab{c}}. Towards
  the science of security and privacy in machine learning. arXiv preprint
  arXiv:1611.03814.

\bibitem[{Pedregosa et~al.(2011)Pedregosa, Varoquaux, Gramfort, Michel,
  Thirion, Grisel, Blondel, Prettenhofer, Weiss, Dubourg, Vanderplas, Passos,
  Cournapeau, Brucher, Perrot, and Duchesnay}]{scikit-learn}
Pedregosa, F., Varoquaux, G., Gramfort, A., Michel, V., Thirion, B., Grisel,
  O., Blondel, M., Prettenhofer, P., Weiss, R., Dubourg, V., Vanderplas, J.,
  Passos, A., Cournapeau, D., Brucher, M., Perrot, M., Duchesnay, E., 2011.
  Scikit-learn: Machine learning in {P}ython. Journal of Machine Learning
  Research 12, 2825--2830.

\bibitem[{Quinlan(1993)}]{quinlan1993c4}
Quinlan, J.~R., 1993. C4. 5: Programming for machine learning. Morgan
  Kauffmann.

\bibitem[{Rndic and Laskov(2014)}]{rndic2014practical}
Rndic, N., Laskov, P., 2014. Practical evasion of a learning-based classifier:
  A case study. In: IEEE Symposium on Security and Privacy (SP). IEEE, pp.
  197--211.

\bibitem[{Rowe et~al.(2007)Rowe, Custy, and Duong}]{rowe2007defending}
Rowe, N.~C., Custy, E.~J., Duong, B.~T., 2007. Defending cyberspace with fake
  honeypots. Journal of Computers 2~(2), 25--36.

\bibitem[{Salem et~al.(2008)Salem, Hershkop, and Stolfo}]{salem2008survey}
Salem, M.~B., Hershkop, S., Stolfo, S.~J., 2008. A survey of insider attack
  detection research. Insider Attack and Cyber Security, 69--90.

\bibitem[{Sculley et~al.(2011)Sculley, Otey, Pohl, Spitznagel, Hainsworth, and
  Zhou}]{sculley2011detecting}
Sculley, D., Otey, M.~E., Pohl, M., Spitznagel, B., Hainsworth, J., Zhou, Y.,
  2011. Detecting adversarial advertisements in the wild. In: Proceedings of
  the 17th ACM SIGKDD international conference on Knowledge discovery and data
  mining. ACM, pp. 274--282.

\bibitem[{Sethi and Kantardzic(2015)}]{sethi2015don}
Sethi, T.~S., Kantardzic, M., 2015. Don’t pay for validation: Detecting
  drifts from unlabeled data using margin density. Procedia Computer Science
  53, 103--112.

\bibitem[{Sethi and Kantardzic(2017{\natexlab{a}})}]{sethi2017data}
Sethi, T.~S., Kantardzic, M., 2017{\natexlab{a}}. Data driven exploratory
  attacks on black box classifiers in adversarial domains. arXiv preprint
  arXiv:1703.07909.

\bibitem[{Sethi and Kantardzic(2017{\natexlab{b}})}]{sethi2017reliable}
Sethi, T.~S., Kantardzic, M., 2017{\natexlab{b}}. On the reliable detection of
  concept drift from streaming unlabeled data. Expert Systems with
  Applications.

\bibitem[{Sethi et~al.(2016)Sethi, Kantardzic, and Hu}]{sethi2016grid}
Sethi, T.~S., Kantardzic, M., Hu, H., 2016. A grid density based framework for
  classifying streaming data in the presence of concept drift. Journal of
  Intelligent Information Systems 46~(1), 179--211.

\bibitem[{Sethi et~al.(2017)Sethi, Kantardzic, and Ryu}]{tsethi2016}
Sethi, T.~S., Kantardzic, M., Ryu, J.~W., 2017. Security theater: On the
  vulnerability of classifiers to exploratory attacks. In: 12th Pacific Asia
  Workshop on Intelligence and Security Informatics. Springer.

\bibitem[{Singh et~al.(2012)Singh, Walenstein, and
  Lakhotia}]{singh2012tracking}
Singh, A., Walenstein, A., Lakhotia, A., 2012. Tracking concept drift in
  malware families. In: Proceedings of the 5th ACM workshop on Security and
  artificial intelligence. ACM, pp. 81--92.

\bibitem[{Smutz and Stavrou(2016)}]{smutz2016tree}
Smutz, C., Stavrou, A., 2016. When a tree falls: Using diversity in ensemble
  classifiers to identify evasion in malware detectors.

\bibitem[{Stein et~al.(2011)Stein, Chen, and Mangla}]{stein2011facebook}
Stein, T., Chen, E., Mangla, K., 2011. Facebook immune system. In: Proceedings
  of the 4th Workshop on Social Network Systems. ACM, p.~8.

\bibitem[{Stevens and Lowd(2013)}]{stevens2013hardness}
Stevens, D., Lowd, D., 2013. On the hardness of evading combinations of linear
  classifiers. In: Proceedings of the 2013 ACM workshop on Artificial
  intelligence and security. ACM, pp. 77--86.

\bibitem[{Tram{\`e}r et~al.(2016)Tram{\`e}r, Zhang, Juels, Reiter, and
  Ristenpart}]{tramer2016stealing}
Tram{\`e}r, F., Zhang, F., Juels, A., Reiter, M.~K., Ristenpart, T., 2016.
  Stealing machine learning models via prediction apis. In: USENIX Security.

\bibitem[{Vorobeychik and Li(2014)}]{vorobeychik2014optimal}
Vorobeychik, Y., Li, B., 2014. Optimal randomized classification in adversarial
  settings. In: Proceedings of the 2014 international conference on Autonomous
  agents and multi-agent systems. International Foundation for Autonomous
  Agents and Multiagent Systems, pp. 485--492.

\bibitem[{Wang(2015)}]{wang2015robust}
Wang, F., 2015. Robust and adversarial data mining.

\bibitem[{Wo{\'z}niak et~al.(2014)Wo{\'z}niak, Gra{\~n}a, and
  Corchado}]{wozniak2014survey}
Wo{\'z}niak, M., Gra{\~n}a, M., Corchado, E., 2014. A survey of multiple
  classifier systems as hybrid systems. Information Fusion 16, 3--17.

\bibitem[{Xu et~al.(2014)Xu, Guo, Zhao, Erbacher, Zhu, and
  Liu}]{xu2014comparing}
Xu, J., Guo, P., Zhao, M., Erbacher, R.~F., Zhu, M., Liu, P., 2014. Comparing
  different moving target defense techniques. In: Proceedings of the First ACM
  Workshop on Moving Target Defense. ACM, pp. 97--107.

\bibitem[{Zhang et~al.(2015)Zhang, Chan, Biggio, Yeung, and
  Roli}]{zhang2015adversarial}
Zhang, F., Chan, P.~P., Biggio, B., Yeung, D.~S., Roli, F., 2015. Adversarial
  feature selection against evasion attacks.

\bibitem[{{\v{Z}}liobait{\.e}(2010)}]{vzliobaite2010learning}
{\v{Z}}liobait{\.e}, I., 2010. Learning under concept drift: an overview. arXiv
  preprint arXiv:1010.4784.

\end{thebibliography}




\end{document}